\documentclass[10pt,journal,compsoc]{IEEEtran}

\usepackage{epsfig}
\usepackage{graphicx}
\usepackage{amsmath}
\usepackage{amssymb}
\usepackage{subcaption}
\usepackage{sidecap}
\usepackage{verbatim}
\usepackage{hyperref}
\usepackage{color}

\begin{document}

\title{A Lightweight Neural Network for Monocular View Generation with Occlusion Handling}

\author{{Simon~Evain,
        and~Christine~Guillemot}
\IEEEcompsocitemizethanks{
\IEEEcompsocthanksitem E-mail: simon.evain@irisa.fr, christine.guillemot@inria.fr
\IEEEcompsocthanksitem  S. Evain and C. Guillemot are with INRIA Rennes Bretagne Atlantique, Campus de Beaulieu, 35042 Rennes, France.}
\thanks{This work has been funded by the EU H2020 Research and Innovation Programme under grant agreement No 694122 (ERC advanced grant CLIM).}}


\IEEEtitleabstractindextext{
\begin{abstract}
In this article, we present a very lightweight neural network architecture, trained on stereo data pairs, which performs view synthesis from one single image. With the growing success of multi-view formats, this problem is indeed increasingly relevant. The network returns a prediction built from disparity estimation, which fills in wrongly predicted regions using a occlusion handling technique. To do so, during training, the network learns to estimate the left-right consistency structural constraint on the pair of stereo input images, to be able to replicate it at test time from one single image. The method is built upon the idea of blending two predictions: a prediction based on disparity estimation, and a prediction based on direct minimization in occluded regions. The network is also able to identify these occluded areas at training and at test time by checking the pixelwise left-right consistency of the produced disparity maps.  At test time, the approach can thus generate a left-side and a right-side view from one input image, as well as a depth map and a pixelwise confidence measure in the prediction. The work outperforms visually and metric-wise state-of-the-art approaches on the challenging KITTI dataset, all while reducing by a very significant order of magnitude (5 or 10 times) the required number of parameters (6.5 M).\end{abstract}

}

\maketitle
\IEEEdisplaynontitleabstractindextext
\IEEEpeerreviewmaketitle

\IEEEraisesectionheading{\section{Introduction}\label{sec:introduction}}

\IEEEPARstart{B}{eing} able to synthesize new viewpoints for a given scene is a classical objective in computer vision, and it has been the subject of intense research over the past two decades. Most approaches for this task (\cite{woodford07}, \cite{penner17}, \cite{flynn15}, \cite{habtegebrial18}) seek to generate those new views from multiple input frames of the scene. By contrast, synthesizing new viewpoints when given only one input image is a more challenging problem, which has garnered less attention from the vision community.

Yet, techniques aiming at generating new views from one single image can have very useful applications. First, they naturally entail a better understanding of the 3D scene from one image, which is crucial for 3D reconstruction. Besides, the past advancements and predicted developments in the coming years of multi-view formats, such as 3D, VR or light field contents, give a significant importance to these techniques. They can indeed be seen as an efficient way to compress these memory-consuming formats. Since more and more of these contents are consumed on mobile devices, it could also prove to be very beneficial for the related methods to be as computationally efficient and lightweight as possible.

Even though research on the subject is not new by any means (\cite{horry97}, \cite{hoiem05}), the emergence of machine-learning based methods in the recent years has dramatically changed the prospects of the field; indeed, due to the significantly ill-posed nature of the problem, they can be especially relevant. In particular, neural networks permitted several major breakthroughs in the field of computer vision in the recent years, and they are thus tools that we prioritize for this kind of problem.

Several recent works have managed to obtain results for monocular view synthesis using deep learning techniques (\cite{xie16}, \cite{cun17}). They often rely on a geometrical estimation of the scene from the given image, through the prediction of the pixelwise flow between the input image and the ground truth image at training time. Most of these methods are very parameter-heavy, have a hard time handling synthesis in tricky areas (occlusions, non-Lambertian surfaces,...), and capturing accurately the global structures of the image. Methods were proposed to complete the geometrical analysis with occlusion processing (\cite{park17}, \cite{liu18}), but most of them either are not able to process natural images yet, or require ground truth occlusion maps in the training set, which makes a generalization to more diverse data elements complicated.

In this article, we present a lightweight architecture able to perform view synthesis with occlusion handling in a stereo context, from one single, unlabelled and unannotated image, beyond state-of-the-art performance. Besides, it only requires a small amount of data for training.  In particular, it is able, at training and at test time, to estimate the disparity map corresponding to the problem at hand, and to evaluate a confidence in its prediction when using the estimated disparity map for the synthesis. Knowing this confidence measure, it is then able to refine the value of the pixels wrongly estimated, with a refinement network. The end result is a prediction built from a geometrical analysis of the scene, which is filled in wrongly predicted areas using a occlusion handling technique. Since 3D scene information is extracted in the course of the analysis, multiple new views can then be generated by interpolation. The architecture is composed of three components, a Disparity-Based Predictor (\textbf{DBP}), a Refiner (\textbf{REF}) and a Confidence-Based Merger (\textbf{CBM}).

We show the efficiency of our approach by notably applying it on the challenging, wide-baseline stereo dataset KITTI (\cite{geiger12}, \cite{geiger13}, \cite{menze15}), with convincing and realistic-looking results for our synthesized images. We show that our method visually and metric-wise outperforms the state-of-the-art methods Deep3D (\cite{xie16}) and \cite{godard17} on stereo view generation, while having far fewer parameters (around 6M). We also show that it is able to perform view synthesis accurately even for scenes with large occluded regions, with no requirement to have any ground truth occlusion map for the training. Besides, it is also scalable, and can be applied efficiently on images of various resolutions. The source code as well as the trained network are publicly available at: \url{http://clim.inria.fr/research/MonocularSynthesis/monocular.html}.

In summary, our contributions are:
\begin{itemize}
    \item An architecture able to outperform state-of-the-art monocular stereo view synthesis approaches, with a number of parameters reduced by an order of magnitude (5 to 10 times when compared with state-of-the-art methods in the field). The training code as well as the network are available at: \url{http://clim.inria.fr/research/MonocularSynthesis/monocular.html}.
    \item A way of handling occlusions in a monocular setting through the learning of forward-backward consistency.
    \item A training schedule in 3 steps, which is key to guiding the output towards a good prediction in spite of the very reduced number of parameters.
    \item A scalable architecture, that can be applied to images of various resolutions, and that can naturally interpolate a set of high-quality views in-between the input view and the stereo predicted one.
\end{itemize}

\section{Related work}

The proposed approach performs depth-based monocular view synthesis with occlusion processing. We briefly review here the main papers of the literature relevant to our work.

\subsection{Monocular depth estimation}
Given the ill-posed nature of the monocular depth estimation problem, learning-based techniques are widely used. In \cite{saxena09}, Saxena et al. managed to obtain interesting results by segmenting the image into superpixels. They then tried to infer the 3D position and orientation of every superpixel using MRFs. The method was an interesting take on the subject, but its main drawback is that since all decisions are made locally, it lacks the global consistency that natural disparity maps should have.

In \cite{eigen14}, Eigen et al. managed to show that neural networks are a good fit for this kind of problem, by designing an architecture able to directly learn disparity from raw pixels. This method was further improved in more recent works (\cite{cao16}, \cite{laina16}). Still, these methods are supervised, and require disparity maps as ground truth elements in the training set. Since disparity maps are not easy to capture for real-life images, synthetic images tailored to a specific set-up are regularly used, which makes the approach hard to generalize.

Accounting for that limit, recent articles have tried to set the problem in an unsupervised setting, where only a pair of unannotated images are given as input. One of the most notable recent examples is the work of Godard et al. (\cite{godard17}) which seeks to generate two disparity maps from each image in the pair, and adds a consistency metrics to guide the final prediction. In \cite{kuznietsov17}, monocular depth estimation is also performed by means of a combination of supervised learning on ground truth disparity maps, and unsupervised learning on pairs of images. If the disparity maps finally obtained in both cases are very good quality, the approaches themselves are not optimized for view synthesis, and are naturally not able to handle occlusions correctly. For stereo datasets with wide baselines such as KITTI, where disparity gaps, and thus occlusion areas, are large, this can drastically change the realism of the final output. Besides, the networks themselves are rather parameter-heavy, and it is hard to imagine them being used on a mobile device. 

\begin{figure*}[!htb]
\centering

\minipage{0.33 \linewidth}
\includegraphics[width=\linewidth]{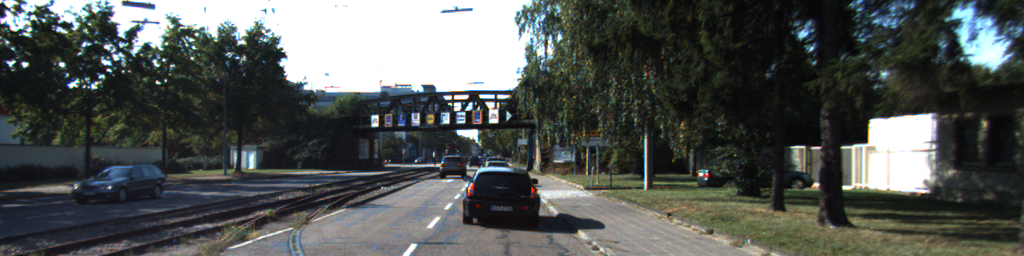}
\includegraphics[width=\linewidth]{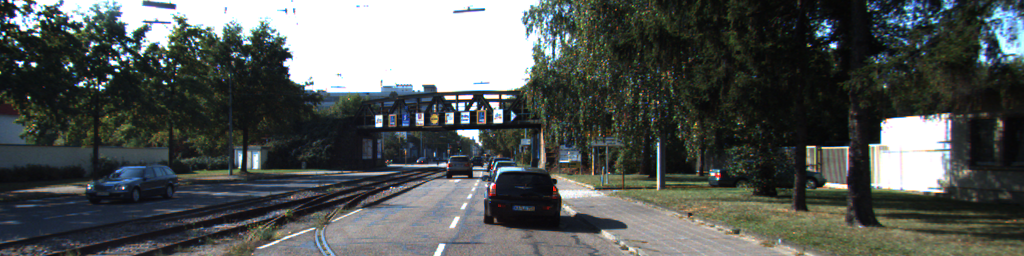}
\includegraphics[width=\linewidth]{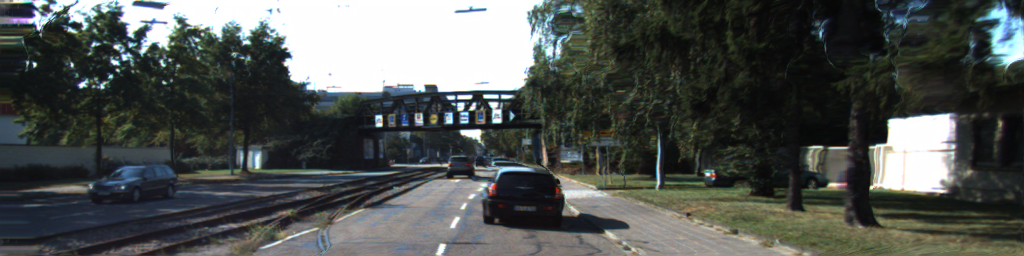}
\includegraphics[width=\linewidth]{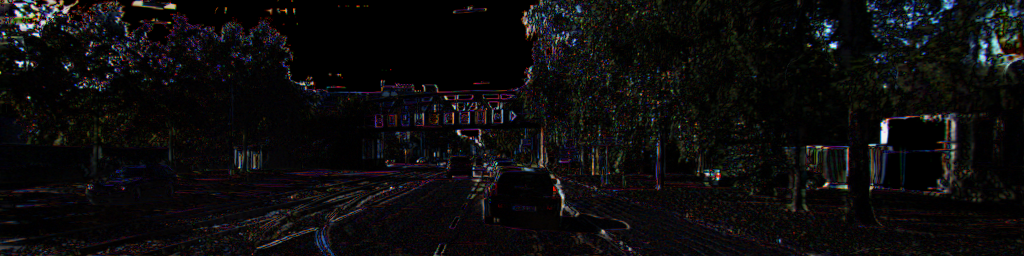}
\includegraphics[width=\linewidth]{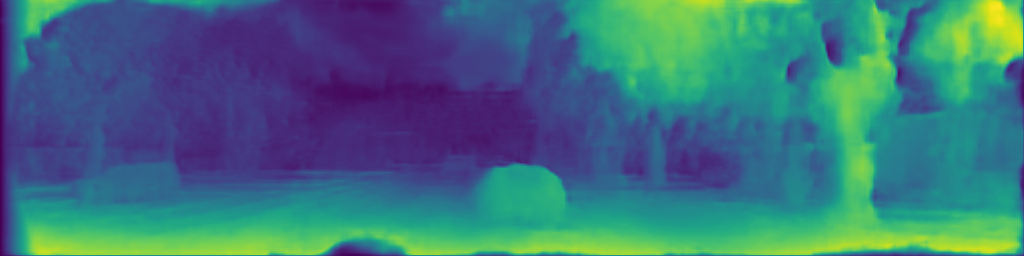}
\endminipage
\minipage{0.33\linewidth}
\includegraphics[width=\linewidth]{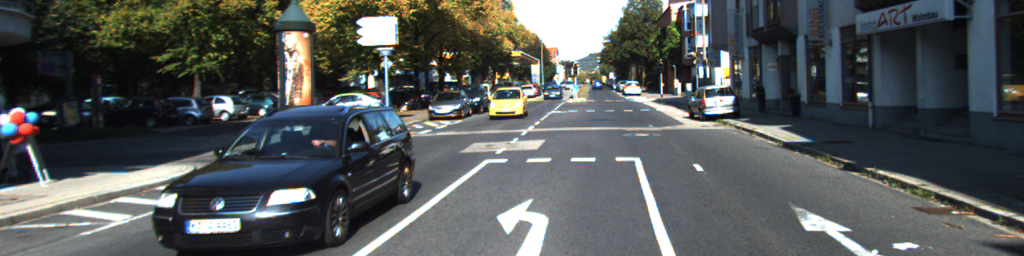}
\includegraphics[width=\linewidth]{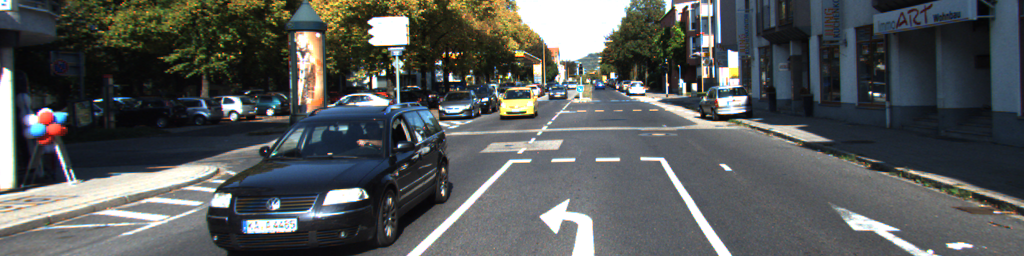}
\includegraphics[width=\linewidth]{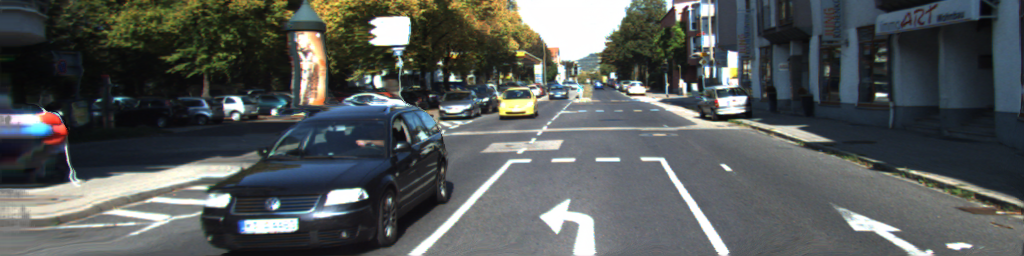}
\includegraphics[width=\linewidth]{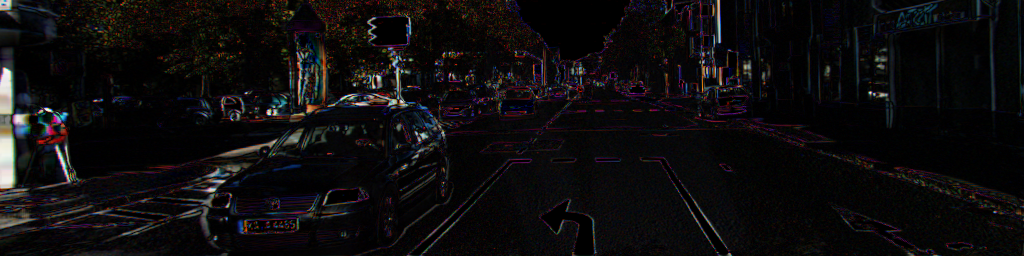}
\includegraphics[width=\linewidth]{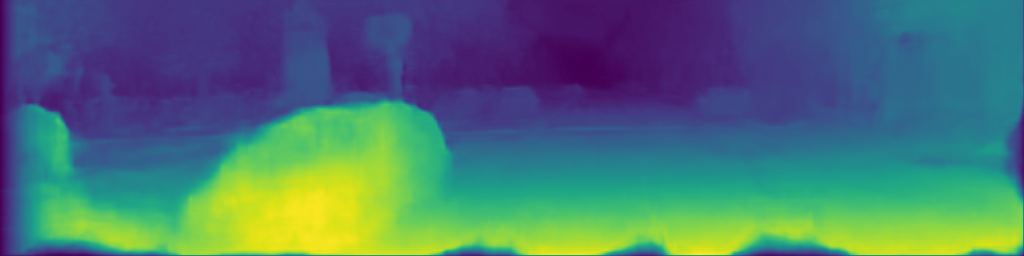}
\endminipage
\minipage{0.33\linewidth}
\includegraphics[width=\linewidth]{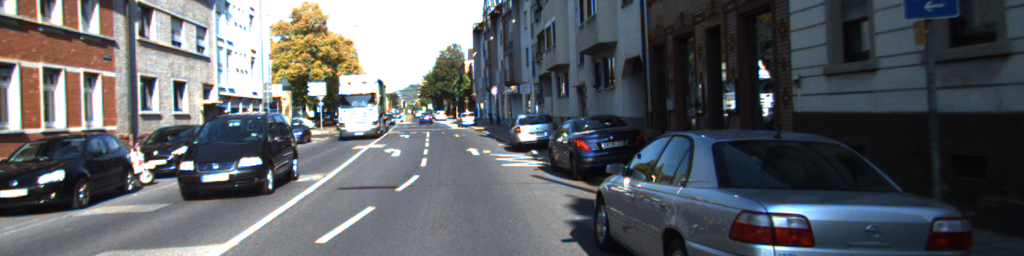}
\includegraphics[width=\linewidth]{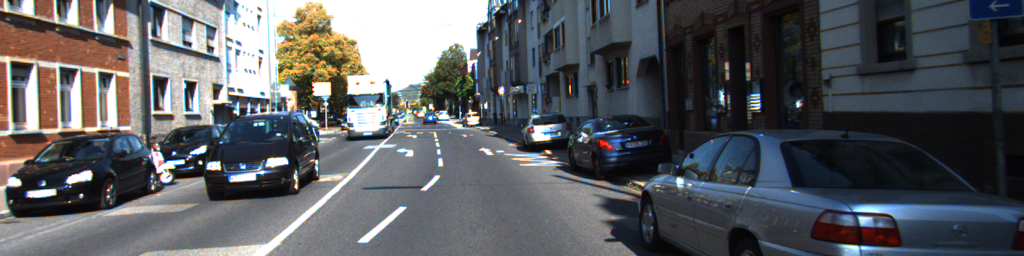}
\includegraphics[width=\linewidth]{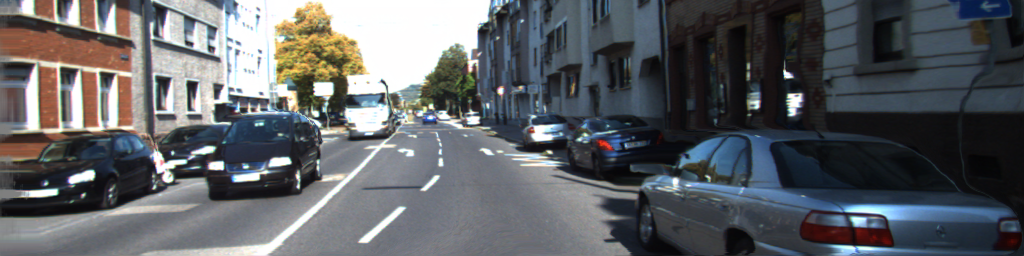}
\includegraphics[width=\linewidth]{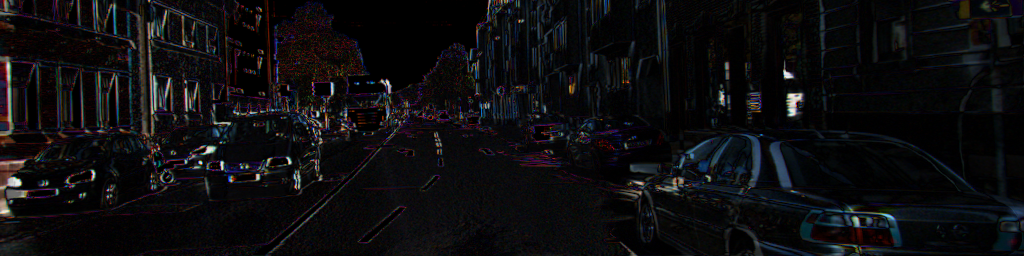}
\includegraphics[width=\linewidth]{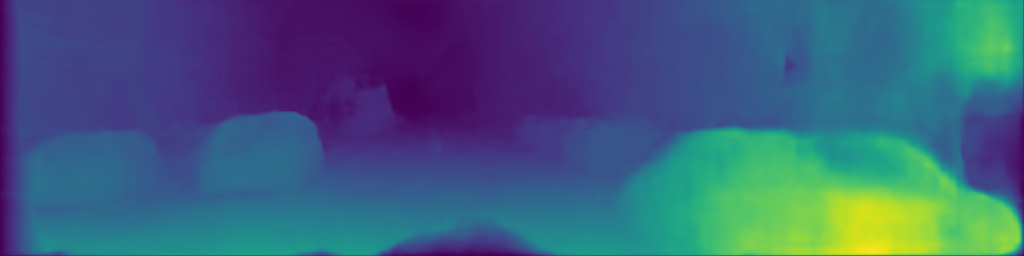}
\endminipage

\caption{Results of our approach on 3 examples from the KITTI test set. From top to bottom: input to the network, ground truth image, prediction carried out by the network, $l1$ error, estimated disparity map.}
\label{ours_result}
\end{figure*}

\subsection{Monocular view synthesis}

Monocular view synthesis is the field of research that aims at synthesizing new viewpoints of a scene given only one input image. Before neural networks became essential tools in computer vision, the methods that were deployed either relied on very clearly defined semantics, and were thus efficient on one very specific type of image only (\cite{hoiem05}), or analyzed the perspective of the input image and used it to find the vanishing point (\cite{horry97}), with interesting results. Still, they were obviously hardly accurate for all images and perspectives.

Due to the ill-posed nature of the problem, strong priors need to be defined: the information needed for synthesizing new viewpoints is indeed not present in the input image. For that reason, machine-learning-based techniques, and in particular deep-learning-based techniques, are powerful tools.

In \cite{kulkarni15}, Kulkarni et al. developed a generative autoencoder model which learns an internal representation of the input image (generated from 3D models of faces in their case). By modifying the hidden variables composing the internal representation, the approach is then able to synthesize new viewpoints. However, it implies that our data elements may be decomposed following a certain set of known variables. This can be assumed for face images, but can hardly generalize well to generic scenes. Besides, the only possible new viewpoints in that case are slightly rotated versions. In \cite{zhou16}, Zhou et al. estimated the motion field between the input image and the ground truth. Even if interesting results are obtained, they are blurry and still suffer from visible artifacts. An iteration over this work has been performed in \cite{sun18}, where the authors develop a method for synthesizing views from one image and a 6DoF vector, by predicting an image from the blending of a flow-based method and a pixel generator method. The available trained network returns impressive results within a specified range of transformations, but is not well-adapted to the stereo setting with a significant disparity gap that is our work setting. Besides, it is very parameter-heavy. To overcome these issues, in \cite{liu18}, Liu et al. build an approach based on homography estimation. Although the results are vastly improved, significant blur and artifacts are still present when working on natural images, notably in the specific stereo case we focus on.

Indeed, we focus, in our work, on one specific case of monocular view generation, based on static stereo contents. In this case, at training time, we have as input a stereo pair, while at test time we want to be able to generate from one single image another stereo view. The reference work in the domain, Deep3D, was designed by Xie et al. in \cite{xie16}. At training time, it takes as input pairs of stereo images, and is able at test time to produce a right-side view whenever an image is sent as input. Learning is carried out through the means of a probabilistic disparity map. This approach has several drawbacks, such as the inability to handle occlusions explicitly and its high number of parameters (around 60 millions for wide baseline and a 256 $\times$ 512 image), especially when the input is high resolution. Besides, the approach is not scalable, since the size of the architecture depends both on the input resolution and the considered disparity range. In \cite{cun17}, Cun et al. also deployed a method for monocular view generation, relying on a pre-trained depth estimator, allowing to obtain good results. Yet, the results presented are obtained on dense multi-view datasets, where disparities are small. In our case, we specifically focus on more complex datasets, where the disparity ranges are significantly higher. In \cite{srinivasan17}, the authors deploy a method for light field generation from one single view, using a 2-stage learning process, estimating geometry first, and then occluded rays. To build its prediction, the method learns the epipolar constraints on light fields to be able to replicate it. This method is a very interesting take on the subject, but is restricted to very simple and similar settings (flower images in this case), and the resort to the epipolar constraint also means that the method cannot be applied on high-disparity, or stereo contents.

\subsection{Occlusion processing}

The problem of processing occlusions when only one image is given is of course an extremely ill-posed problem, since it implies being able to return information that is unavailable at test time. Even though in most cases, the occlusions are not explicitly considered in the learning process, several recent approaches have tried to address this problem. 

In \cite{park17}, Park et al. use a specific encoder-decoder network to handle the occluded areas. In \cite{liu18}, Liu et al. use a similar idea, but slightly modify the loss function used for learning. These approaches are interesting, but they require a significant number of parameters. Besides, the occluded areas are not identified automatically by the network, but instead given as ground truth occlusion maps. This implies, for the learning process, an easy access to ground truth occlusion maps, which complicates the resort to the algorithm when working on natural images. Unlike these methods, our approach is able to estimate the occlusion map within the learning process with no required ground truth.

Finally, Tulsiani et al., in \cite{tulsiani18}, also predict the disoccluded pixels from the images that they produce. To do so, they use a DispNet architecture (around 40M parameters) and a 2 layer-based view synthesis process, to capture both the visible points and the occluded regions. The method that we present in this article is lighter, and captures the disoccluded pixels directly by focusing on the disparity estimation inconsistencies.

\section{Notations}
\hspace*{0.40cm}
* $L$, $R$: left and right ground truth images.

* $L_{DBP}, R_{DBP}$: left and right DBP-based predictions.

* $L_{REF}, R_{REF}$: left and right REF-based predictions.

* $L^{*}, R^{*}$: left and right final predictions.

* $d_{LR}, d_{RL}$: estimated disparity map for left-to-right (respectively right-to-left) view synthesis.

* $C_{LR}, C_{RL}$: confidence depth maps for left-to-right (respectively right-to-left) view synthesis as obtained by the network from the disparity estimations, only computed during training.

* $V_{LR}, V_{RL}$: estimated confidence depth maps for left-to-right (respectively right-to-left) view synthesis, as estimated by the network from $C_{LR}$ and $C_{RL}$, so as to be used at test time.

\section{Our method}

\begin{figure*}
\centering
\includegraphics[width = 15cm]{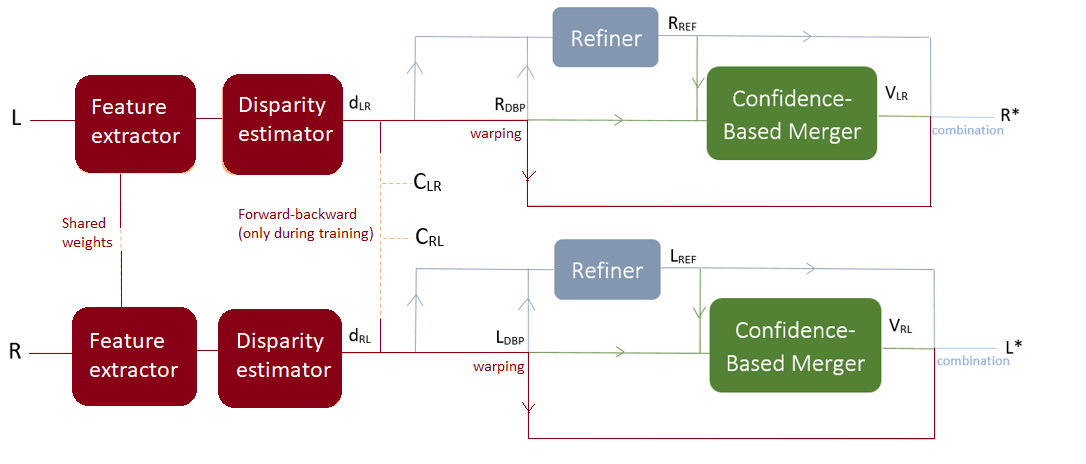}
\caption{Graph of the overall structure of our approach.  Dark red blocks represent DBP, the blue block represents REF, and the green block CBM.}
\label{overall_structure}
\end{figure*}

In this paper, we introduce an end-to-end differentiable approach for monocular view generation, able to synthesize new viewpoints from one single image. The work is performed in a stereo setting, meaning that the training dataset is made up of stereo pairs, with a significant disparity gap between them. Before delving into the in-depth description of every component, and into the way the learning proceeds, let us first focus on the overall structure of our approach, which is also depicted in figure \ref{overall_structure}. The exact architecture of every component (with number of filters, kernel sizes, strides for every single layer enumerated) is shown on \url{http://clim.inria.fr/research/MonocularSynthesis/monocular.html}.

\subsection{Overall structure}

The architecture can be decomposed into 3 main component networks (more details are given in their specific sections):

The \textbf{Disparity-based Predictor (DBP)}, which seeks to estimate the disparity map between the two images at training time, by learning to warp one image from the pair onto the other viewpoint. This gives us a first prediction based on disparity map estimation. Yet, this prediction is not able to handle occlusions, and is prone to errors since the global structure of the image can hardly be captured by this pixelwise estimation.

The \textbf{Refiner (REF)} seeks to enhance the DBP-based prediction by means of filtering. Since the main objective of this network is to cope with the flaws of the disparity-based prediction, the intuition is that it will be  important for the areas that cannot be matched by disparity and for which significant information is missing at test time, such as occluded areas. The role of this network is essentially to provide a realistic and plausible output in these regions.

The \textbf{Confidence-Based Merger (CBM)} learns the best way to combine the two complementary predictions obtained by DBP and REF, in order to obtain a good-quality final view. 

Since many of these elements are actually interrelated, the learning schedule is key to guaranteeing the stability and efficiency of the approach.

\subsection{Disparity-based Predictor (DBP)}

First, we want to learn how to generate a disparity map from one given image, and then use it to predict the new view by warping the input view. We want to make sure that at test time, it can be done automatically using only one input image.

To do so, we consider a convolutional neural network, in which the last layer is a spatial transformer layer (\cite{jaderberg15}).

\subsubsection{Learning architecture}

The learning architecture is built as an encoder-decoder structure, with skip-connections so that no information is lost in the downsampling part of the encoder. The intuition behind this architecture is to consider the encoder part of the network as a feature extractor from the input, and the decoder part as a section which processes these features to generate the actual disparity map.

The encoder is made of a MobileNet 1.0 architecture (\cite{howard17}), where the last layers devoted to classification have been removed. The MobileNet networks are a class of lightweight neural networks which, despite their low number of parameters, are able to compete with most state-of-the-art approaches in image classification. They are characterized by the replacement of standard convolution filters within the network with a succession of depthwise convolutional filters and 1 $\times$ 1 pointwise convolution filters. This allows to greatly reduce the number of parameters at hand, all while maintaining a high number of feature maps. The architecture is made up of 13 successive -convolution 1 $\times$ 1, depthwise convolution 3 $\times$ 3- blocks with a gradually increasing number of filters at every block. We also initialize the weights of the encoder with pre-trained weights on ImageNet (\cite{deng09}). Initializing the encoder with pre-trained weights mirrors the fact that this part of the architecture is devoted to feature extraction. We practically found that it brought major improvements when compared with random initialization.

We then simply design the decoder as a symmetrical counterpart to the encoder, with 5 blocks, where every block is in this case -depthwise convolution 3 $\times$ 3, convolution 1 $\times$ 1, upsampling of 2, skip-connection-. At the end of the decoder, the network returns a matrix with the same resolution as the input image, and we want this matrix to be an estimation of the disparity map of our input image. The last layer of the DBP is then a spatial transformer layer (\cite{jaderberg15}), which, similarly to \cite{liu17} and \cite{amersfoort17}, has no trainable parameters and uses the output of the learning architecture as a motion field to warp the input image and form the prediction. The network can then be trained directly on the images, and generate the disparity as an intermediate result. 

\subsubsection{Overall learning structure}

The key idea of DBP is to use as much data as possible at training time. Since we have a pair of images accessible during training, we thus learn to perform a left-to-right, and a right-to-left view synthesis at the same time, with two independent branches (see figure \ref{overall_structure}). We consider that the feature extraction process is common to both tasks ; we share the weights of the encoder process in the two branches. However, we train independently the decoder (disparity estimator) in both branches. This gives us as output an access to two disparity maps $d_{LR}$ and $d_{RL}$ which have been independently trained from each other on their respective branches. The point of having these two disparity maps learned separately is that it will allow us to check their consistency and evaluate the confidence that we may have in our prediction for every pixel (see section 3.4). Besides, sharing the encoder weights between the two branches is also helpful since it means that the feature extractor will be fed with twice as many data elements as it would have if only one branch existed.

At test time, any of the two branches can be used individually, allowing us, depending on the chosen path, to generate a left-side or right-side view from any input image. The component, including both branches, contains around 6M parameters.

\subsubsection{Limits of DBP}

This component gives us a first prediction, based on the estimated disparity map. Now, a prediction entirely built on disparity maps has inherent flaws.

Indeed, the areas that are occluded in the input image have to be filled in the synthesized view, and the way to inpaint those pixels cannot be given by a sole disparity map. The non-Lambertian surfaces, and more globally the differences in lighting can make the matching process difficult. Besides, the performed pixelwise prediction may suffer from a loss in spatial coherence. A small error on the position of the candidate pixel can also lead to very visible artifacts.

For those reasons, we add a new component network, the \textbf{Refiner} (\textbf{REF}) which takes as input the prediction synthesized by DBP, and which has for objective to fix the issues listed above.

\subsection{Refiner (REF)}

Before we describe the architecture of REF, let us discuss the design philosophy of it.

\subsubsection{Optimizing for a direct error metrics}

We want to be able to post-process the regions where the disparity-based prediction fails. For many of these pixels, the information is actually unavailable at test time (notably in occluded areas). At this stage, we will then, in these regions, directly perform a $l1$ minimization. The network will seek to remove the artifacts and the errors produced, by refining them with neighbouring pixels. The end result runs the risk of being blurry, and of losing some important details of the scene. For that reason, we only want the Refiner to operate in areas where the DBP is not sufficiently accurate. The detection of these regions where DBP fails will be done thanks to a confidence map estimated by the third component CBM. 

\subsubsection{Architecture of REF}

REF is designed as a very simple architecture made up of a succession of 8 convolutional layers, all of them (except for the last one) having 64 3 $\times$ 3 filters (see \url{http://clim.inria.fr/research/MonocularSynthesis/monocular.html} for more details).

\subsection{Confidence-Based Merger (CBM)}

The DBP and REF based predictions are both imperfect, but complementary. Indeed, DBP retains all details from the input image, but presents very strong and unpleasant artifacts when the matching is not accurate. Notably, disocclusions cannot be handled by this component. On the opposite side, REF produces an image with less artifacts, but details are lost in the process. The main objective of the CBM is to be able to combine these two predictions into one optimized final prediction. To do so, we want to be able to estimate a pixelwise confidence measure in our DBP. Indeed, if for one pixel we have a high confidence in our DBP, we will prefer the DBP pixel, since it carries more details. Conversely, if the confidence is low, the REF pixel will be preferred, with fewer visible artifacts. This will help us improve the result of our approach in occluded regions.

\subsubsection{Confidence measure - Identification}

To define this confidence measure on the DBP, we use the fact that at training time we have two estimated disparity maps: from left-to-right ($d_{LR}$) and right-to-left ($d_{RL}$) view synthesis. The following forward-backward consistency relations can be defined:
\begin{equation}
\begin{split}
d_{RL}(x,y) &= d_{LR}(x-d_{RL}(x,y), y) \\
d_{LR}(x,y) &= d_{RL}(x+d_{LR}(x, y), y)
\end{split}
\end{equation}
The confidence measure is built to check whether the relations are verified for every pixel of the two disparity maps. If so, there is a consistency between the two predictions. Otherwise, it means that there was a problem in the disparity estimation process for this pixel. These relations lead us to define two confidence maps (one per branch), where $\gamma$ is a parameter controlling the decay rate of the confidence measure function of the warping error :
\begin{equation}
\begin{split}
C_{RL}(x,y) &= \exp(-\gamma |d_{RL}(x,y) - d_{LR}(x-d_{RL}(x,y), y)|) \\
C_{LR}(x,y) &= \exp(-\gamma |d_{LR}(x,y) - d_{RL}(x+d_{LR}(x, y), y)|)
\end{split}
\end{equation}
This way, if the relations are verified, the value for the corresponding estimated confidence will be close to 1. Conversely, if they are not, the confidence values will tend to get closer to 0.

\subsubsection{Final synthesis - Combination}

This confidence measure is available at training time, because we have access to the two images, and thus the two disparities, but it cannot be used as such at test time. For that reason, a third part of the network (see figure \ref{overall_structure}) is devoted to learning the overall appearance of these confidence maps, from one prediction only. The architecture for learning this map is made up of 5 successive convolutional layers with (except for the last one) 32 3 $\times$ 3 filters (see \url{http://clim.inria.fr/research/MonocularSynthesis/monocular.html} for more details). It should be noted that we do not expect our approximation of the confidence maps to seek for the exact same values, but instead to be able to discriminate low-confidence from high-confidence pixels.

Considering the notations from section 3, the final predictions $L^{*}$ and $R^{*}$ can be written as:
\begin{equation}
\begin{split}
L^{*} &= V_{RL}L_{REF}+(1-V_{RL})L_{DBP}\\ 
R^{*} &= V_{LR}R_{REF}+(1-V_{LR})R_{DBP}\\ 
\end{split}
\end{equation}
where $V_{RL}$ and $V_{LR}$ are respectively the estimations of $(1-C_{RL})$ and $(1-C_{LR})$ carried out by CBM. Since $V_{LR}$ and $V_{RL}$ are initialized with values close to 0, it allows us to have as a starting point, for our final prediction, $L_{DBP}$ and $R_{DBP}$. This way, we pick pixels from the disparity-based prediction when the confidence is high, and from REF when it is low. At test time, choosing either one of the two branches allows to produce a left-side or right-side view from any input image. The last activation function of the CBM is sharp, leading $V_{LR}$ and $V_{LR}$ to values very close to 0 or 1 ; this way we will tend to reduce the blurriness of the final result.

\section{Learning process}

Many of the components presented in the previous section are obviously interrelated, and thus a joint learning of all these components would risk to be unstable and inefficient. For that reason, a specific learning schedule needs to be specified to optimize its performance.

\subsection{Phase I: DBP}

As a starting point, we only learn the disparity-based prediction. Using the notations of section 3, we define the learning metrics as:
\begin{equation}
\begin{split}
& \lambda_0 (||L_{DBP}-L||_1 + ||R_{DBP}-R||_1)\\
& + \lambda_1 (||\nabla L_{DBP} - \nabla L||_1 +  ||\nabla R_{DBP} - \nabla R||_1)
\end{split}
\end{equation}
We choose the $l^1$ metrics, following notably the analysis carried out in \cite{mathieu16}. We jointly train the two branches, and in order to better capture the structure of the image, we add a gradient-based loss.

\subsection{Phase II: Geometrical restructuring of DBP}

To make sure that the estimated disparity map captures with as much accuracy as possible the various structures of the input image, we add a regularization step (drawing inspiration from \cite{zimmer11}). In other words, we use the following learning metrics:
\begin{equation}
\begin{split}
& \lambda_2 (||\frac{2}{\max(d_{RL})} \nabla d_{RL} - \nabla L||_1 \\
& + ||\frac{2}{\max(d_{LR})} \nabla d_{LR} - \nabla R||_1 )\\
& + \lambda_3 (||L_{DBP}-L||_1 + ||R_{DBP}-R||_1)
\end{split}
\end{equation}
We constrain the normalized (to keep its value between -2 and 2) gradient of our disparity maps to be as close as possible to the gradient of our ground truth images, in order to better capture the various structures of the image. Besides, we retain a pixelwise term in the learning metrics to make sure that the prediction remains close to the ground truth element. Unlike many works (\cite{liu17}, \cite{amersfoort17}), we do not resort to a multi-scale approach to tackle the geometrical structuring, for the sake of reducing the number of parameters of the network.

\subsection{Phase III: REF and CBM}

We finally focus on the REF and CBM pipelines. In this last step, we freeze the weights of DBP. We do it because we do not want the whole process to interfere with the quality of the disparity maps that were produced so far. Besides, the first two steps allow to generate two disparity maps, which can then be used to generate corresponding confidence maps. By freezing the learning weights for disparity, we make sure that the confidence measure is fixed, making its estimation possible and stable.

We use the following learning metrics for our final prediction:
\begin{equation}
\begin{split}
& \lambda_4 (||L_{REF}-L||_1 + ||R_{REF}-R||_1)\\
&+ \lambda_5 (||\nabla L_{REF} - \nabla L||_1 +
||\nabla R_{REF} - \nabla R||_1)\\
&+ \lambda_6 (||L^{*}-L||_1 + ||R^{*}-R||_1)\\
&+ \lambda_7 (||\nabla L^{*} - \nabla L||_1 + ||\nabla R^{*} - \nabla R||_1)\\
&+ \lambda_8 (||V_{LR} - (1-C_{LR})||_1 + ||V_{RL}-(1-C_{RL})||_1)
\end{split}
\end{equation}
In the end, we obtain estimated confidence maps ($V_{LR}$ and $V_{RL}$), as well as the final predictions $L^{*}$ and $R^{*}$.

\section{Experiments}

\begin{figure*}[!htb]
\centering

\minipage{0.32\linewidth}
\centering
\begin{subfigure}{0.02\linewidth}
\caption{}
\end{subfigure}

\includegraphics[width=\linewidth]{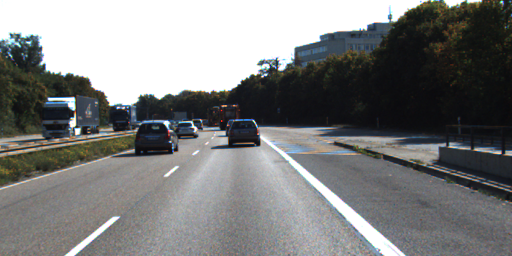}
\includegraphics[width=\linewidth]{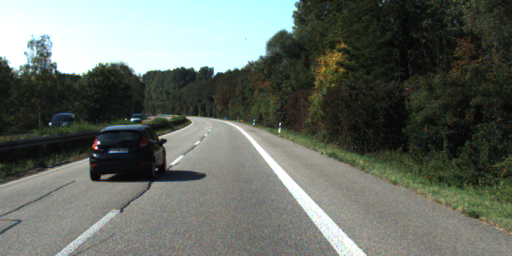}
\includegraphics[width=\linewidth]{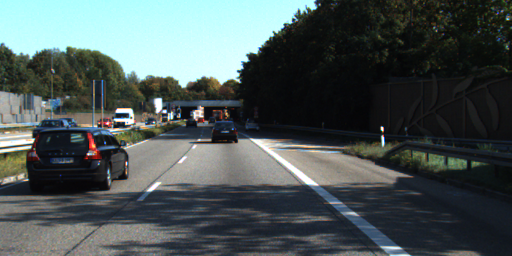}
\endminipage
\minipage{0.32\linewidth}
\centering
\begin{subfigure}{0.02\linewidth}
\caption{}
\end{subfigure}
\includegraphics[width=\linewidth]{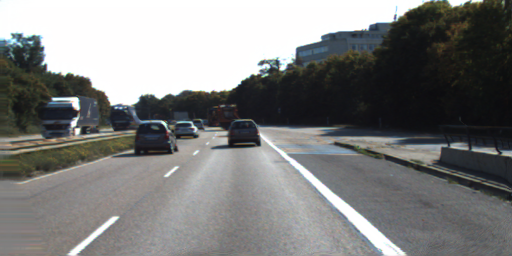}
\includegraphics[width=\linewidth]{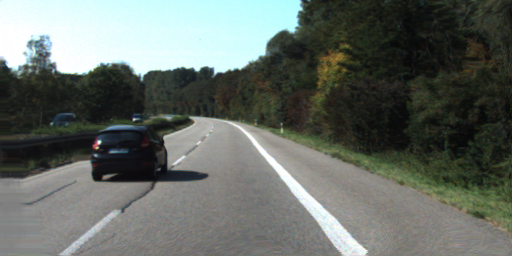}
\includegraphics[width=\linewidth]{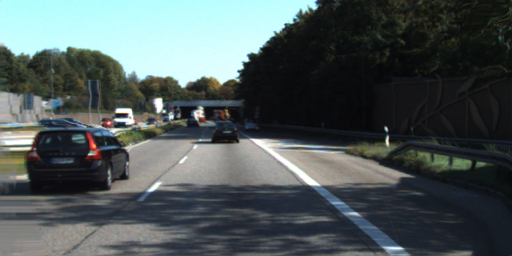}
\endminipage
\minipage{0.32\linewidth}
\centering
\begin{subfigure}{0.02\linewidth}
\caption{}
\end{subfigure}
\includegraphics[width=\linewidth]{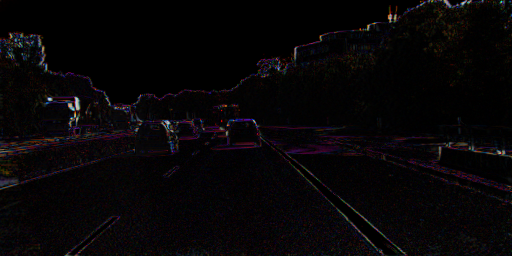}
\includegraphics[width=\linewidth]{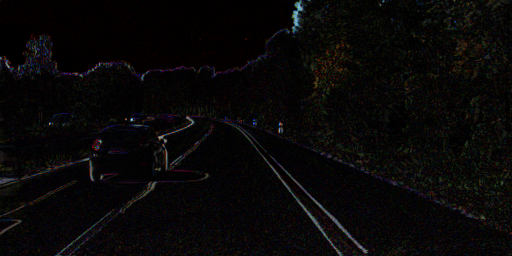}
\includegraphics[width=\linewidth]{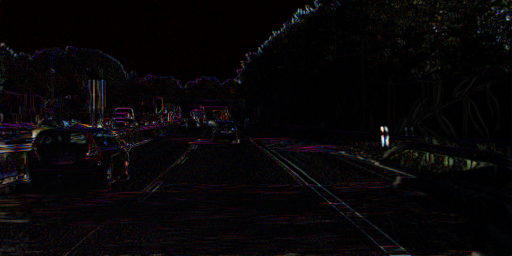}
\endminipage

\caption{Qualitative evaluation of the predictions carried out. a) Ground truth image. b) Our prediction. c) L1 error between the prediction and the ground truth image.}
\label{eval_gt_pred}
\end{figure*}

In this section, we show the results, strengths and limits of our model. In order to evaluate its efficacy, we perform the comparison on stereo datasets with wide baselines, mostly in the context of automatic driving. The results are thus evaluated metric-wise and visually on the KITTI dataset (\cite{geiger12}, \cite{geiger13}, \cite{menze15}). Visual results are presented in figures \ref{ours_result} and \ref{eval_gt_pred}. We also advise the reader to check \url{http://clim.inria.fr/research/MonocularSynthesis/monocular.html}, which displays more numerous and more diverse high-resolution examples of comparisons.

\subsection{Implementation}

Before feeding them into the network, following the preprocessing steps from \cite{liu17}, we normalize all the images into a $[-1,1]$ range. During training, we extract patches (with a $256 \times 256$ resolution) from the pair of images as input. We also perform color data augmentation on-the-fly randomly for 20 \% of the input elements, with random gamma and brightness transformations. Our model is trained with a batch size of 16 using Adam  (\cite{kingma15}) as the optimization algorithm, with $\beta_1 = 0.9$ and $\beta_2 = 0.999$.

The network, which was implemented in TensorFlow (\cite{abadi15}) and Keras (\cite{chollet15}), has around 6.5 million parameters as a whole, and takes only a few hours to be fully trained on a Tesla P100 GPU. The learning rate is chosen as $0.0001$, and is halved when there is no improvement after 10 epochs. The learning is stopped when the validation metrics has not improved after 20 epochs. When unspecified, all weights are initialized following a random normal distribution. We opt for the following values for our hyperparameters: $\gamma = 0.07$, $\lambda_0 = 0.80$, $\lambda_1=0.20$, $\lambda_2 = 0.85$, $\lambda_3 = 0.15$, $\lambda_4 = 0.25$, $\lambda_5 = 0.05$, $\lambda_6 = 0.50$, $\lambda_7 = 0.13$, $\lambda_8 = 0.035$.

\subsection{Evaluation}
\label{evaluation}
We evaluate the quality of our method using different metrics. First, we consider PSNR as a reference measure. PSNR allows to measure the pixelwise error between the prediction and the ground truth image (the higher, the better). It is indeed a canonical measure, but it has its flaws: for one, it is not really able to measure the structural reconstruction quality of the prediction, for every pixel is considered independently from its neighbors. Besides, it can not evaluate the perceptual quality of the image produced, since a small offset of a few pixels in the prediction can drastically reduce the PSNR score, all while having a perceptual impact close to none.

To address the two precited drawbacks, we decide to use two more metrics. We resort to SSIM, since it is a metrics (the higher, the better) that is more fitting to evaluate the structural quality of the prediction, and is thus an interesting complement to PSNR evaluation. We also use LPIPS (\cite{zhang18}), which is a deep feature-based distance well suited for evaluating the 'perceptiveness' of our prediction (the lower, the better). Following the analysis in \cite{zhang18}, we specifically choose the Alex-lin network for evaluation. Combining these three metrics for our prediction is a good way to have a full comparison between the various methods.

Finally, we want to evaluate the quality of our method specifically on disoccluded regions to quantify the contribution brought by our occlusion handling component. To identify these regions on the unannotated KITTI test set, following the protocol from \cite{tulsiani18}, we use an off-the-shelf stereo matching algorithm (\cite{yamaguchi14}). We consider the disoccluded pixels as the pixels that could not be matched with this method. We then compute the pixelwise error only on those pixels. This allows us to evaluate the performance of the method on regions that are tricky to predict.

To evaluate our approach, we compare it with 2 reference methods: Deep3D (\cite{xie16}) and Godard et al.'s approach from \cite{godard17}. Deep3D aims at producing automatically a right-side image from a left-side image. Godard et al.'s work is focused on monocular disparity estimation, and we want to show that the disparity maps that we produce are better suited for warping than the output of the method that is deployed in \cite{godard17}.

\subsection{Statistical results}
\begin{table}
\centering
\begin{tabular}{|c|c|c|c|c|}
  \hline
  \textbf{KITTI Test set} & \textbf{PSNR} & \textbf{SSIM} & \textbf{LPIPS} & \textbf{params}\\
  \hline
  Ours & \textbf{19.24} & \textbf{0.74} & \textbf{0.139} & \textbf{6.5M}\\
  \hline
  Deep3D (\cite{xie16}) & 19.08 & \textbf{0.74} & 0.220 & 61M\\
  \hline
  Godard et al. (\cite{godard17}) & 18.44 & 0.71 & 0.148 & 30M\\
  \hline
 \end{tabular}
\caption{Statistical evaluations. The higher the PSNR and SSIM, the better. The lower the LPIPS (\cite{zhang18}), the better.} 
\label{tablePSNR}
\end{table}
To highlight the lightweight aspect of our network, we train our network using only the 400 pairs of frames from the training KITTI 2012 and 2015 stereo challenges. The KITTI dataset is a stereo automatic driving dataset with wide baselines. Among these 400 pairs of images, 35 are kept as validation. For evaluation, we use the 400 images from the test sets of the challenge, and perform the evaluations when working with a right-side image as input, and a left-side image to be predicted. Since Deep3D needs to be trained for a specific input resolution and disparity range, and Godard et al.'s available network is optimized for 512 $\times$ 256 images, we decide to center-crop the images from the KITTI test set so as to obtain 512 $\times$ 256 images for evaluation. The evaluation in the case of Godard et al. is the average over the pixels that were actually predicted, when using the warped disparity map predicted by the method. The statistical results are displayed in table \ref{tablePSNR}. We can see that we obtain slightly better results in terms of PSNR and SSIM than Deep3D (and clearly above Godard et al.'s approach).

We notice that our approach significantly outperforms the other methods in terms of perceptual quality. In particular, we note that Deep3D, while close in PSNR, is very distant when looking at this perceptual metrics. We conclude that with a number of parameters very clearly lower than these state-of-the-art methods, we manage to outperform them both in PSNR/SSIM, and significantly in more perceptual measures such as LPIPS.

\subsection{Visual comparison on KITTI}

\begin{figure*}[!htb]
\centering

\minipage{0.32\linewidth}
\centering
\begin{subfigure}{0.02\linewidth}
\caption{}
\end{subfigure}

\includegraphics[width=\linewidth]{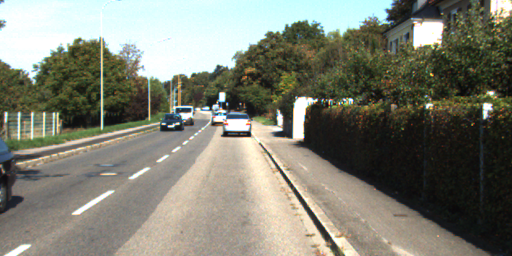}
\includegraphics[width=\linewidth]{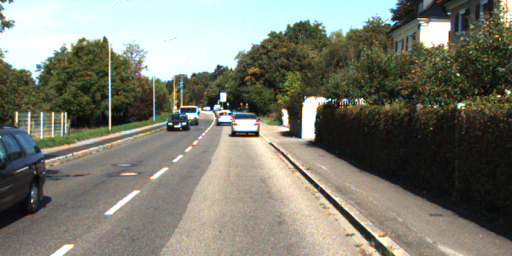}
\includegraphics[width=\linewidth]{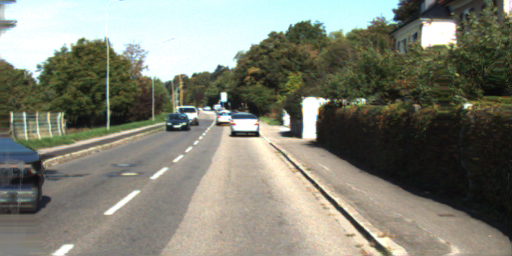}
\includegraphics[width=\linewidth]{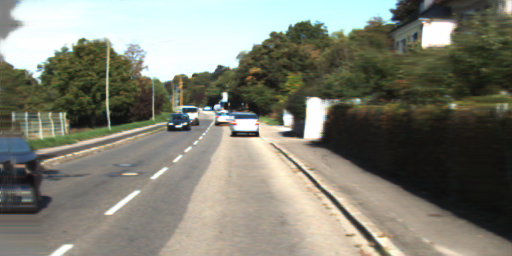}
\includegraphics[width=\linewidth]{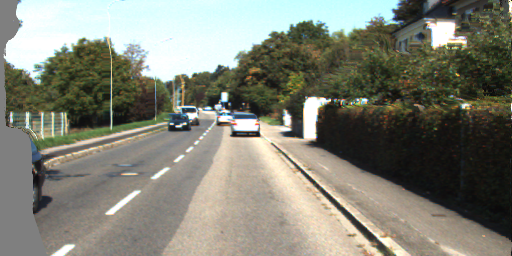}
\endminipage
\minipage{0.32\linewidth}
\centering
\begin{subfigure}{0.02\linewidth}
\caption{}
\end{subfigure}
\includegraphics[width=\linewidth]{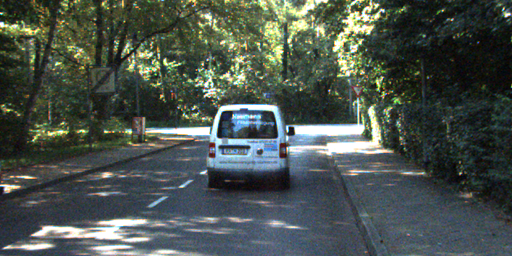}
\includegraphics[width=\linewidth]{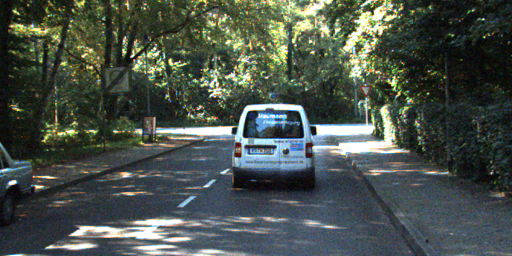}
\includegraphics[width=\linewidth]{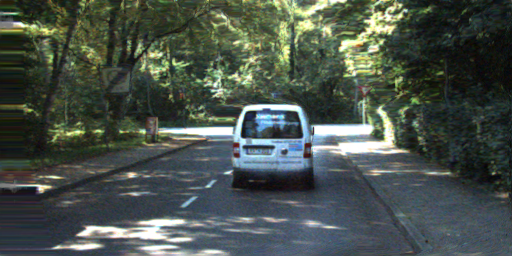}
\includegraphics[width=\linewidth]{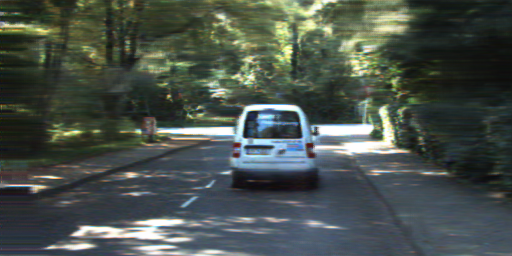}
\includegraphics[width=\linewidth]{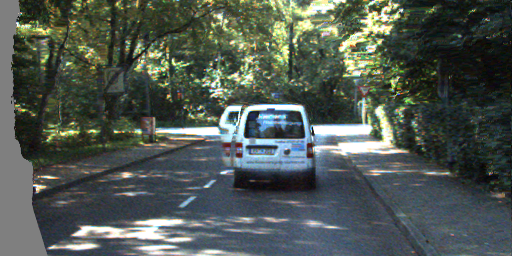}
\endminipage
\minipage{0.32\linewidth}
\centering
\begin{subfigure}{0.02\linewidth}
\caption{}
\end{subfigure}
\includegraphics[width=\linewidth]{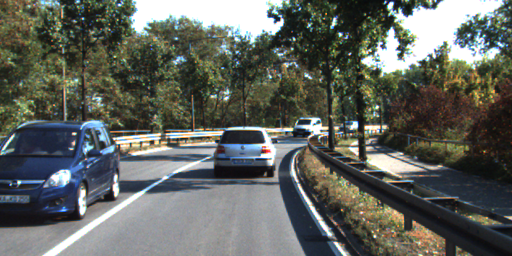}
\includegraphics[width=\linewidth]{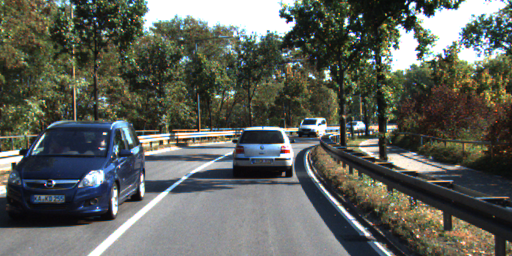}
\includegraphics[width=\linewidth]{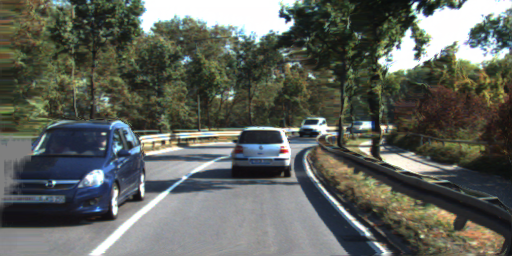}
\includegraphics[width=\linewidth]{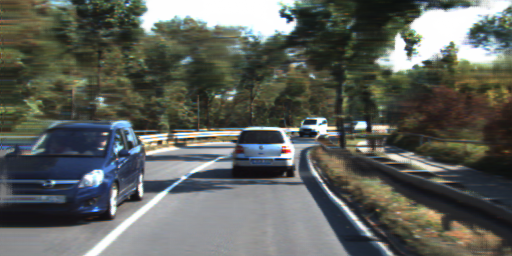}
\includegraphics[width=\linewidth]{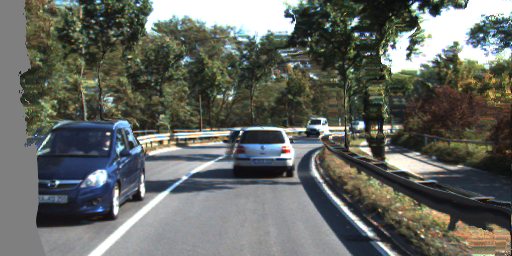}
\endminipage

\caption{Comparison of the approaches on three examples from the KITTI test set (from top to bottom: input image, ground truth image, our method, Deep3D (\cite{xie16}) and Godard et al. (\cite{godard17})).}
\label{comparisons_kitti}
\end{figure*}

The visual difference in terms of quality of our method with the other techniques is much more significant than the numerical difference. A good way to evaluate the quality of the algorithm is thus to look at the generated images themselves, and see how realistic they are. Some results on the KITTI test set are shown in figure \ref{comparisons_kitti}. 

We note that visually, the images produced by Deep3D are usually more blurry and have less accurate details than in our approach. This blurring can have a pronounced effect, and it leads to the fading of certain structures (such as the sign in the second image from figure \ref{comparisons_kitti}). Our approach is much sharper, and thus provides a more realistic and plausible appearance to our predicted images.

When compared with Godard et al.'s approach, we note that our method is better at handling structures and disoccluded pixels. This is particularly notable when looking at the disoccluded regions from the cars, which are not handled correctly by Godard's approach. Besides, we note that the trees in the third image are also not processed accurately by the algorithm. Generally speaking, we note that our approach is better at processing structures within the image and handling disoccluded regions.

\begin{figure}[!htb]
\centering

\minipage{0.14\linewidth}
\includegraphics[width=\linewidth]{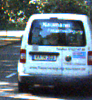}
\includegraphics[width=\linewidth]{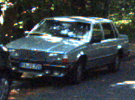}

\endminipage
\minipage{0.14\linewidth}
\includegraphics[width=\linewidth]{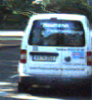}
\includegraphics[width=\linewidth]{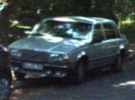}

\endminipage
\minipage{0.14\linewidth}%
\includegraphics[width=\linewidth]{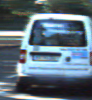}
\includegraphics[width=\linewidth]{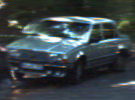}

\endminipage
\minipage{0.14\linewidth}%
\includegraphics[width=\linewidth]{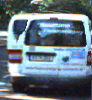}
\includegraphics[width=\linewidth]{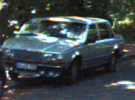}

\endminipage

\caption{Details from KITTI views, (from left to right:) ground truth detail, detail from our prediction, detail from Deep3D, detail from Godard et al.}
\label{comparisons_zoom_kitti}
\end{figure}

\subsection{Ablation study}

\subsubsection{Contribution of the various steps of the training schedule}

\begin{figure}[!htb]
\minipage{\linewidth}
\includegraphics[width=0.32\linewidth]{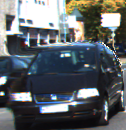}
\includegraphics[width=0.32\linewidth]{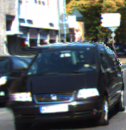}
\includegraphics[width=0.32\linewidth]{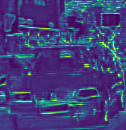}
\endminipage
\caption{Detail from views to illustrate the contribution of the Refiner. From left to right: prediction from DBP, final prediction, confidence map.}
\label{dbp_contribution}
\end{figure}

Now, let us perform an ablation study of our approach, so as to show the benefits of using the various components of our method, and this specific training schedule. To perform these evaluations, we use the metrics defined in section \ref{evaluation}.

The quantitative evaluations of the various components are displayed in table \ref{tablePSNRphases}. Looking at PSNR and SSIM, we would be inclined to think that training the network end-to-end or setting no constraint on the confidence map would be the best way forward. In a way, since these two approaches are directly optimized for a pixelwise metrics, with no deviating constraint (the training schedule or the structure of the blending weights), this should come as no surprise. Yet, when taking a look at at the LPIPS metrics, which accounts for the perceptual quality of the output image produced, we note that doing this actually contributes to a significant degradation of our image. We conclude from this analysis that the training schedule, defined in the article, is the best possible course of action to obtain good results from a perceptual viewpoint. 

Now, taking a look at the occlusion-related metrics, we can note that understandably, adding phase II in the training schedule does not lead to a significant improvement to handle occluded regions. On the contrary, we can see that phase III brings very significant improvements in these tricky regions, which tends to validate the positive contribution of the REF.

\begin{figure*}[!htb]
\centering

\minipage{0.166\linewidth}

\centering
\begin{subfigure}{0.02\linewidth}
\caption{}
\end{subfigure}
\includegraphics[width=\linewidth]{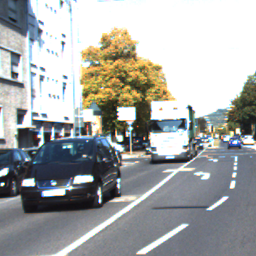}
\includegraphics[width=\linewidth]{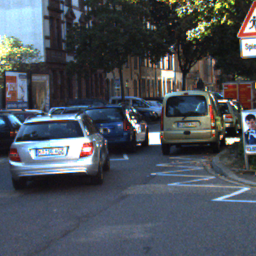}
\includegraphics[width=\linewidth]{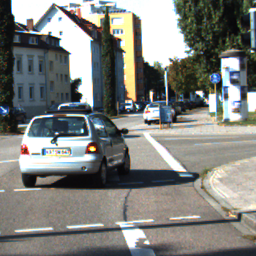}

\endminipage
\hfill
\minipage{0.166\linewidth}
\centering
\begin{subfigure}{0.02\linewidth}
\caption{}
\end{subfigure}
\includegraphics[width=\linewidth]{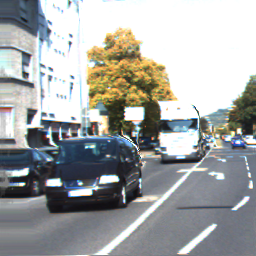}
\includegraphics[width=\linewidth]{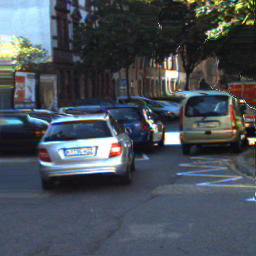}
\includegraphics[width=\linewidth]{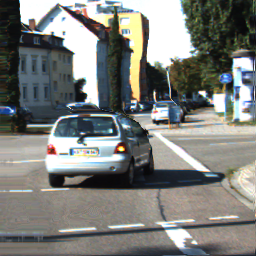}

\endminipage
\hfill
\minipage{0.166\linewidth}
\centering
\begin{subfigure}{0.02\linewidth}
\caption{}
\end{subfigure}
\includegraphics[width=\linewidth]{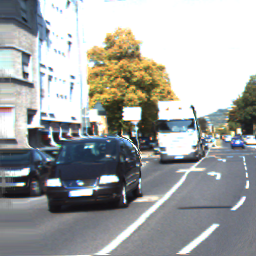}
\includegraphics[width=\linewidth]{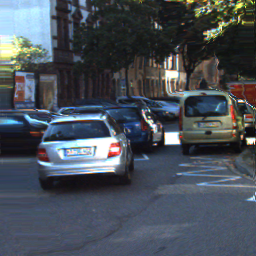}
\includegraphics[width=\linewidth]{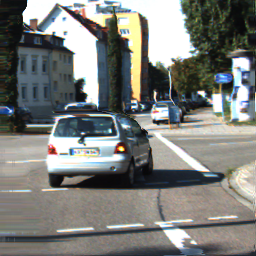}

\endminipage
\hfill
\minipage{0.166\linewidth}
\centering
\begin{subfigure}{0.02\linewidth}
\caption{}
\end{subfigure}
\includegraphics[width=\linewidth]{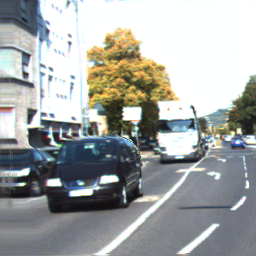}
\includegraphics[width=\linewidth]{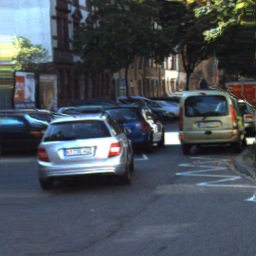}
\includegraphics[width=\linewidth]{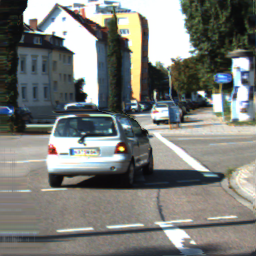}


\endminipage
\minipage{0.166\linewidth}
\centering
\begin{subfigure}{0.02\linewidth}
\caption{}
\end{subfigure}
\includegraphics[width=\linewidth]{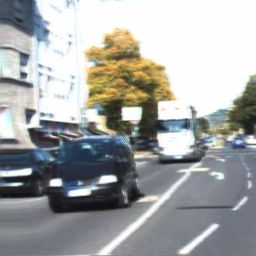}
\includegraphics[width=\linewidth]{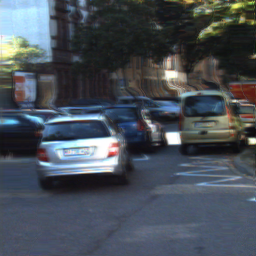}
\includegraphics[width=\linewidth]{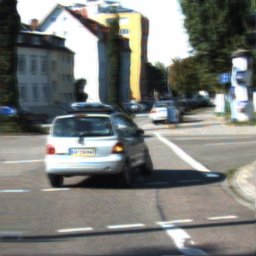}


\endminipage
\minipage{0.166\linewidth}
\centering
\begin{subfigure}{0.02\linewidth}
\caption{}
\end{subfigure}
\includegraphics[width=\linewidth]{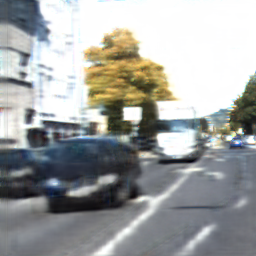}
\includegraphics[width=\linewidth]{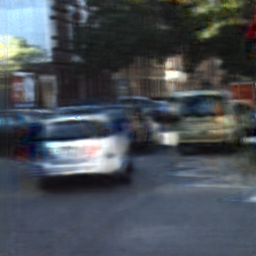}
\includegraphics[width=\linewidth]{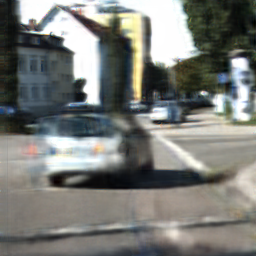}


\endminipage
\caption{Elements of comparison for the ablation study. In each column: \textbf{a)} Input image. \textbf{b)} Result from phase I. \textbf{c)} Result from phases I then II. \textbf{d)} Result from phases I, II then III. \textbf{e)} Result from phases I then III (with phase II skipped). \textbf{f)} Result when trained end-to-end using the metrics from phase III.}
\label{ablation_study_phases}
\end{figure*}

We can now take a look at several of the images that are produced by our algorithm (see figure \ref{ablation_study_phases}).

We can outline several elements:

\begin{enumerate}
    \item Comparing columns \textbf{b} and \textbf{c}, we note at several occasions that phase II indeed improves by a significant margin the structural appearance of the produced images. This is particularly noteworthy when looking at the white truck from row 1, or the yellow building in the background from row 3.
    \item Comparing columns \textbf{c} and \textbf{d}, we note that phase III improves significantly the way occlusions are handled ; this is shown in the artifacts around the foreground car in row 1, or the artifacts from the rightmost car in row 2, which are fixed by the phase III of training. This can also be noticed when looking at figure \ref{dbp_contribution}, which zooms onto an occluded region (around the car) where artifacts are removed by the process.
    \item Looking at column \textbf{e}, we note that skipping phase II usually produces results which are more blurry and less structurally sound. This is understandable by the fact that since the REF operates on a prediction which is far less accurate, it will tend to have a very strong effect to correct the flaws. We see clearly from these images that phase II is an essential component to our learning process, for by improving the quality of the intermediate prediction, it helps REF to be applied to relevant areas only.
    \item Looking at column \textbf{f}, we can clearly understand the advantages of using our training schedule over an end-to-end learning process. Although in terms of PSNR and SSIM, the end-to-end output is very close to the result based on our own training schedule (see table \ref{tablePSNRphases}), we can see that visually, the difference is very significant: our training schedule allows us to obtain a result which is less blurry and far more accurate. By forcing the training process to follow a certain schedule, we thus make sure that our result remains convincing from a perceptual viewpoint.
\end{enumerate}

\begin{table}
\centering
\begin{tabular}{|c|c|c|c|c|c|}
  \hline
  \textbf{KITTI} & \textbf{PSNR} & \textbf{SSIM} & \textbf{LPIPS} & \textbf{PSNR disocc.}\\
  \hline
  Phase I & 18.76 & 0.72 & 0.144 & 14.84\\
  \hline
  Phases I-II & 18.87 & 0.72 & 0.144 & 14.85\\
  \hline
  Phases I-II-III & 19.24 & 0.74 & \textbf{0.139} & 15.32\\
  \hline
  Phases I-III & 19.11 & 0.73 & 0.206 & 15.04\\
  \hline
  Phase III & 19.23 & 0.74 & 0.345 & 15.35\\
  \hline
  No confidence & 19.40 & 0.75 & 0.190 & 15.48\\
  \hline
 \end{tabular}
\caption{Statistical justification of the training schedule. The higher the PSNR and SSIM, the better. The lower the LPIPS (\cite{zhang18}), the better. PSNR disocc accounts for the PSNR on the disoccluded pixel regions only. The comparisons are carried out between networks that have been trained for the mentioned phases of the training schedule. 'Phase III' shows the case where the network is fully trained end-to-end with the metrics from phase III. 'No confidence' shows the result when, during phase III of the training schedule, the metrics constraining the blending weights to be based on the consistency of the disparity maps is removed.} 
\label{tablePSNRphases}
\end{table}

\subsubsection{Constraining the confidence map}

\begin{figure*}[!htb]
\centering
\minipage{0.22\linewidth}

\centering
\begin{subfigure}{0.02\linewidth}
\caption{}
\end{subfigure}
\includegraphics[width=\linewidth]{img/44_p3.png}
\includegraphics[width=\linewidth]{img/27_p3.png}

\endminipage
\hfill
\minipage{0.22\linewidth}
\centering
\begin{subfigure}{0.02\linewidth}
\caption{}
\end{subfigure}
\includegraphics[width=\linewidth]{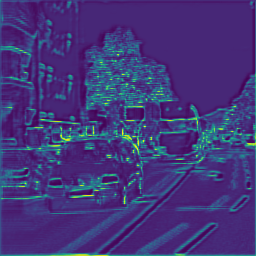}
\includegraphics[width=\linewidth]{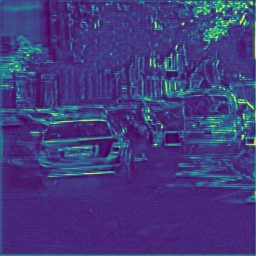}

\endminipage
\hfill
\minipage{0.22\linewidth}
\centering
\begin{subfigure}{0.02\linewidth}
\caption{}
\end{subfigure}
\includegraphics[width=\linewidth]{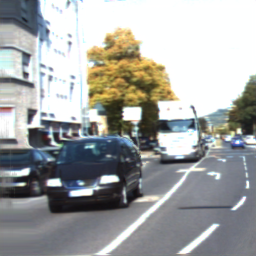}
\includegraphics[width=\linewidth]{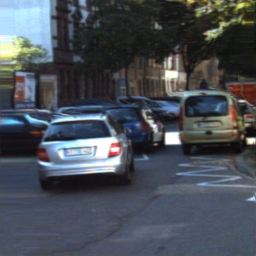}

\endminipage
\hfill
\minipage{0.22\linewidth}
\centering
\begin{subfigure}{0.02\linewidth}
\caption{}
\end{subfigure}
\includegraphics[width=\linewidth]{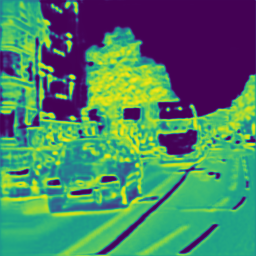}
\includegraphics[width=\linewidth]{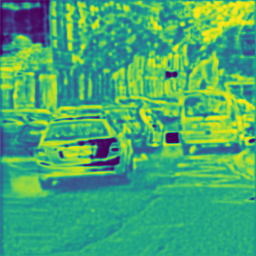}


\endminipage

\caption{Elements of comparison for constraint set on the confidence map (yellow in confidence maps means low-confidence in DBP prediction). \textbf{a)} Prediction when FB-constraint is set. \textbf{b)} Corresponding confidence map. \textbf{c)} Prediction when no FB-constraint is set. \textbf{d)} Corresponding confidence map.}
\label{constraint_confidence}
\end{figure*}

We now compare the results that we obtain when setting a forward-backward confidence constraint in the definition of our blending weights, and when we set no constraint. The comparisons are performed in figure \ref{constraint_confidence}.

We see that the images that we end up obtaining when we do not specify any constraint over the confidence map are usually much more blurry, which is confirmed by the significant difference in terms of LPIPS shown in table \ref{tablePSNRphases}. Besides, figure \ref{constraint_confidence} shows that confidence maps are good representations of occluded or non-Lambertian regions, and that by removing the constraint, we also lose this valuable information.

\subsection{Interpolation process}

One of the interesting features of the approach is that it is not only able to produce stereo views, but that it can also generate a sequence of good-quality interpolated views between the input image and the prediction.
\\

\begin{figure}[!htb]
\includegraphics[width=\linewidth]{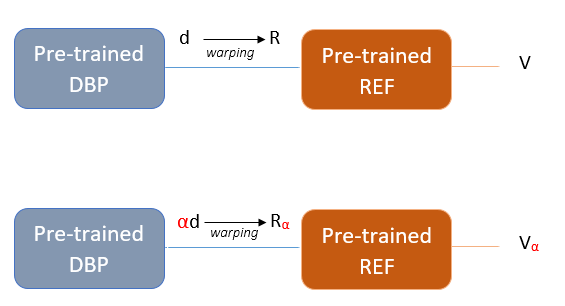}
\caption{Diagram showing the interpolation process. We scale the disparity map obtained as an output of the pre-trained DBP. Warping the input view with the scaled disparity map yields a first interpolated view $R_{\alpha}$. The pre-trained Refiner (and the rest of the network) will then deal with the artifacts, and a confidence map will be generated, so as to produce the final interpolation $V_{\alpha}$ with good quality.}
\label{intepr}
\end{figure}

\begin{figure*}[!htb]
\centering
\minipage{0.32\linewidth}

\centering
\begin{subfigure}{0.02\linewidth}
\end{subfigure}
\includegraphics[width=\linewidth]{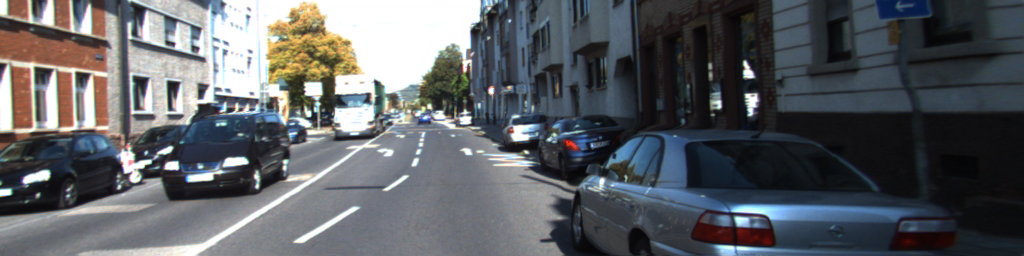}
\includegraphics[width=\linewidth]{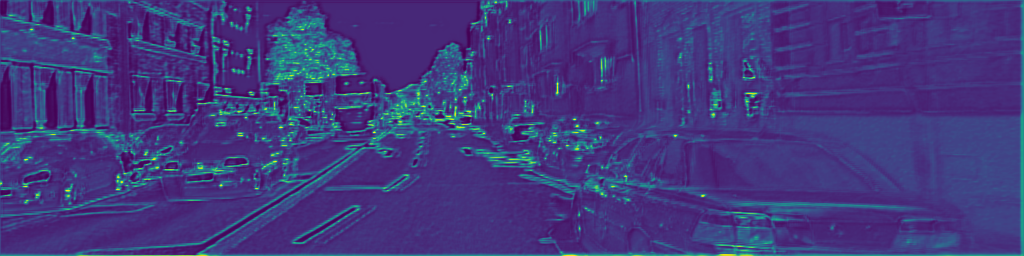}

\endminipage
\hfill
\minipage{0.32\linewidth}
\centering
\begin{subfigure}{0.02\linewidth}
\end{subfigure}
\includegraphics[width=\linewidth]{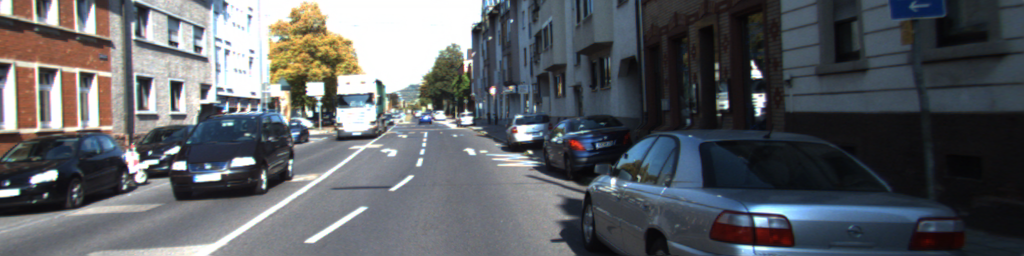}
\includegraphics[width=\linewidth]{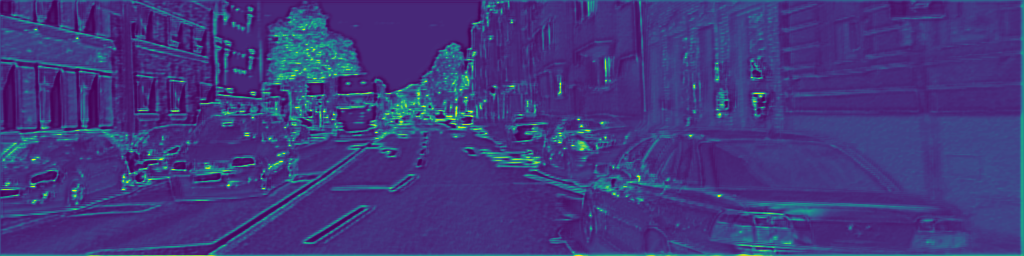}

\endminipage
\hfill
\minipage{0.32\linewidth}
\centering
\begin{subfigure}{0.02\linewidth}
\end{subfigure}
\includegraphics[width=\linewidth]{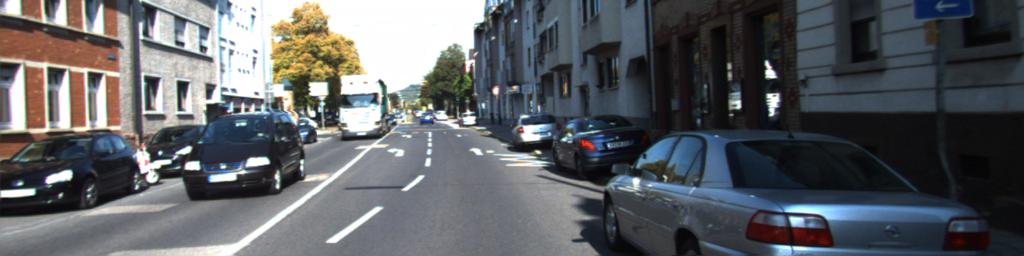}
\includegraphics[width=\linewidth]{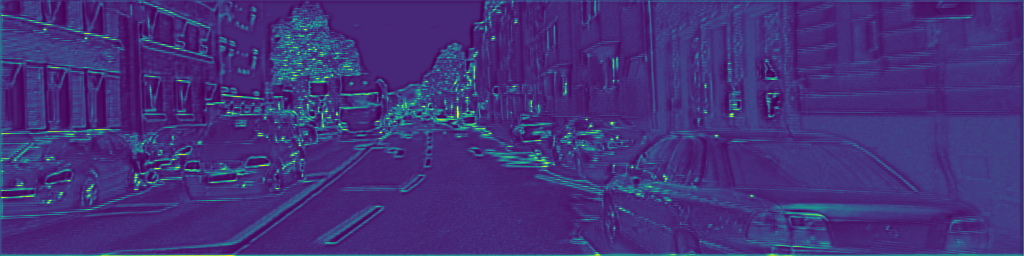}

\endminipage
\caption{Interpolation results (top row) and confidence maps.}
\label{interpolation_views}
\end{figure*}

To do so, we use an already trained network. We scale the disparity map that is obtained as output of the DBP. We use the scaled disparity map to compute by warping a first approximation of the interpolated view. Using the rest of our pre-trained pipeline on the warped views allows us to obtain a good-quality sequence of views around the input image.
\\

\begin{table}[]
  \centering
  \begin{tabular}{|c|c|c|c|c|c|c|}
  \hline
  \textbf{} & \textbf{V0} & \textbf{V1} & \textbf{V2} & \textbf{V4} & \textbf{V5} & \textbf{V6}\\
  \hline
  Ours (PSNR) & \textbf{37.05} & \textbf{39.55} & \textbf{43.39} &
  \textbf{43.36} & \textbf{39.45} & \textbf{36.94}\\
  (SSIM) & \textbf{0.96} & \textbf{0.97} & \textbf{0.99} & \textbf{0.99} & \textbf{0.97} & \textbf{0.96}\\
  \hline
  [23] (PSNR) & 33.74 & 35.65 & 41.04 & 40.87 & 36.58 & 34.25 \\
  (SSIM) & 0.92 & 0.94 & 0.98 & 0.98 & 0.94 & 0.92 \\
  \hline
\end{tabular}
\caption{Metric-wise comparisons between our method and [23] on the Flowers test set. Vi represents the i-th view in the central line of the light field. V3 is the central view (so, the input), and is thus not evaluated. V0 and V6 are the target views, while all the other ones are obtained through our interpolation process. We note that our approach clearly outperforms the work from [23], metric-wise, on all views, even for the interpolated views.}
\label{table_compare_flowers}
\end{table}

Visual results of this interpolation process are shown in figure \ref{interpolation_views}, as well as the corresponding confidence maps built for every interpolated image, when working with large baseline stereo sets, such as KITTI. Visual examples are also shown in the supplementary video on: \url{http://clim.inria.fr/research/MonocularSynthesis/monocular.html}.
\\

In addition, we evaluate quantitatively this interpolation process by working on light field content (the smaller-baseline Flowers dataset, introduced in \cite{srinivasan17}). We train our network on the Flowers training set, by considering stereo pairs (either 'leftmost view - central view' or 'central view' - 'rightmost view') on the central line of the light field. We then evaluate our approach on all the views from this central line, by performing our interpolation process. This way, we are able to evaluate visually and metric-wise the quality of the images that we produce when the central view of the light field is used as input. We compare our results with \cite{srinivasan17}, a method generating a full 4D light field from one single image, using the code provided by the authors. The comparisons are only performed on the central line, and are displayed in table \ref{table_compare_flowers}. We note that our approach clearly outperforms \cite{srinivasan17} on these interpolated views. This shows that our interpolation process is efficient in producing good-quality interpolated views. Besides, it also shows that our approach is able to work efficiently on various datasets, with various baselines and semantics.
\\





\subsection{Results on other datasets}

\begin{figure*}[!htb]
\centering
\minipage{0.24\linewidth}
\centering
\begin{subfigure}{0.02\linewidth}
\caption{}
\end{subfigure}
\includegraphics[width=\linewidth]{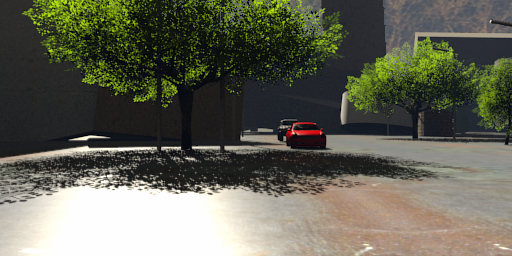}
\includegraphics[width=\linewidth]{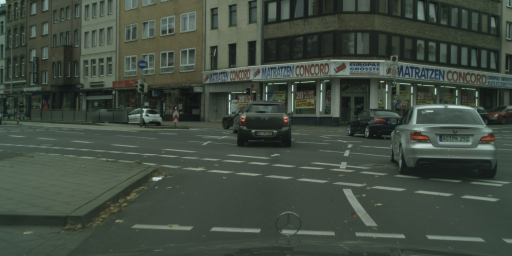}
\includegraphics[width=\linewidth]{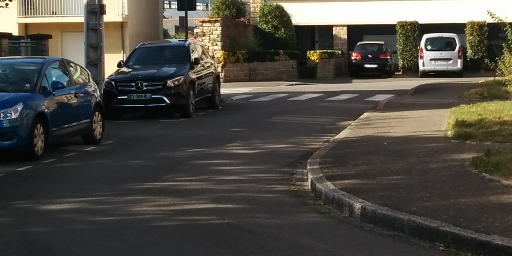}
\endminipage
\hfill
\minipage{0.24\linewidth}
\centering
\begin{subfigure}{0.02\linewidth}
\caption{}
\end{subfigure}
\includegraphics[width=\linewidth]{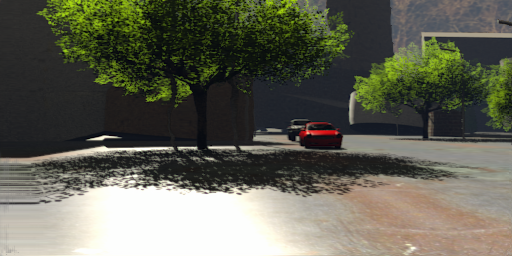}
\includegraphics[width=\linewidth]{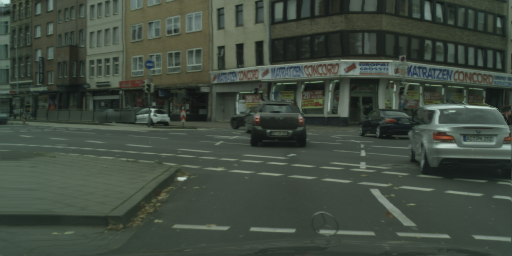}
\includegraphics[width=\linewidth]{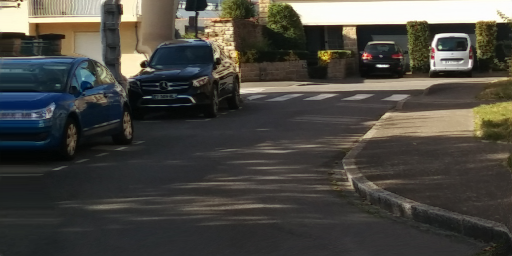}
\endminipage
\hfill
\minipage{0.24\linewidth}
\centering
\begin{subfigure}{0.02\linewidth}
\caption{}
\end{subfigure}
\includegraphics[width=\linewidth]{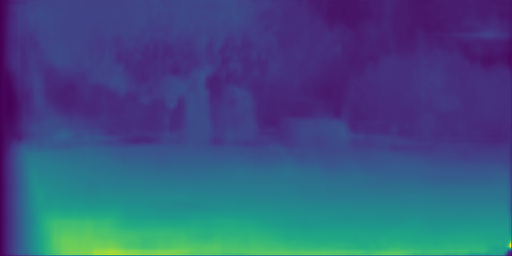}
\includegraphics[width=\linewidth]{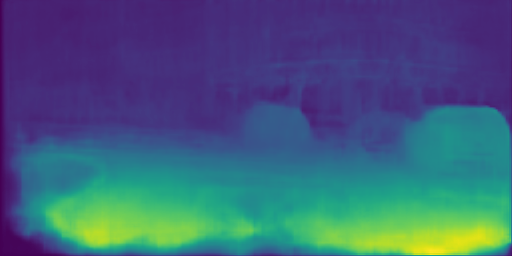}
\includegraphics[width=\linewidth]{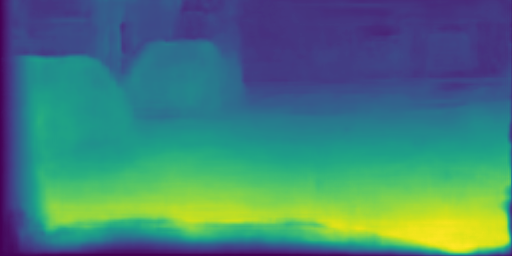}
\endminipage
\hfill
\minipage{0.24\linewidth}
\centering
\begin{subfigure}{0.02\linewidth}
\caption{}
\end{subfigure}
\includegraphics[width=\linewidth]{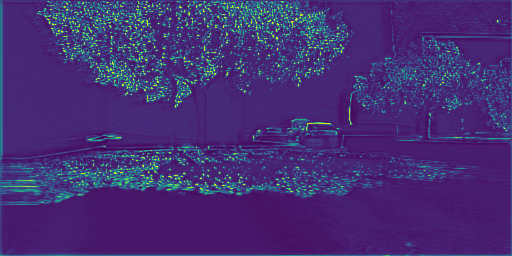}
\includegraphics[width=\linewidth]{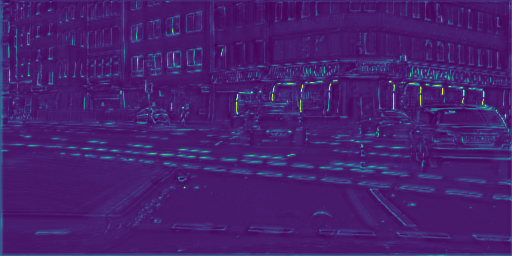}
\includegraphics[width=\linewidth]{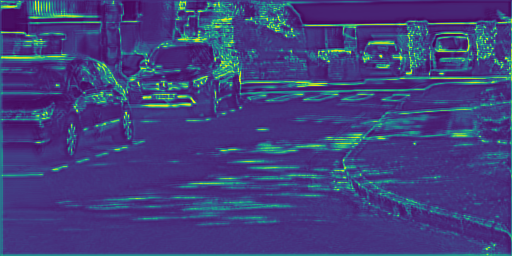}
\endminipage

\caption{Results when the approach, still trained on KITTI, is applied on other urban scenes datasets (from top to bottom: Driving (\cite{mayer16}), Cityscapes (\cite{cordts16}) and pictures taken in Rennes). \textbf{a)} Input image. \textbf{b)} Network prediction. \textbf{c)} Estimated disparity map. \textbf{d)} Estimated confidence map (yellow means low-confidence).}
\label{other_datasets}
\end{figure*}

The network has been trained on the KITTI training set and evaluated on the KITTI test set, but it can be applied efficiently on any kind of images from urban scenes when trained on KITTI. This is shown in figure \ref{other_datasets}.

We can indeed see that our approach, while trained on KITTI only, is able to return plausible disparity maps and to synthesize convincing new views from a synthetic dataset (Driving \cite{mayer16}, first row), another automatic driving dataset (Cityscapes \cite{cordts16}, row 2) and from images captured in natural conditions using a smartphone (Rennes, row 3).
\\

The network was also trained on other datasets, with variable baselines and semantics (such as the small-baseline light field dataset Flowers \cite{srinivasan17} and the 3D movie database Hollywood \cite{hadfield13}), with convincing visual predictions. Visual results on these datasets are shown in figure \ref{other_datasets_hollyflowers} and in the supplementary video (\url{http://clim.inria.fr/research/MonocularSynthesis/monocular.html}).

\begin{figure*}[!htb]
\centering
\minipage{0.32\linewidth}
\centering
\begin{subfigure}{0.02\linewidth}
\end{subfigure}
\includegraphics[width=\linewidth]{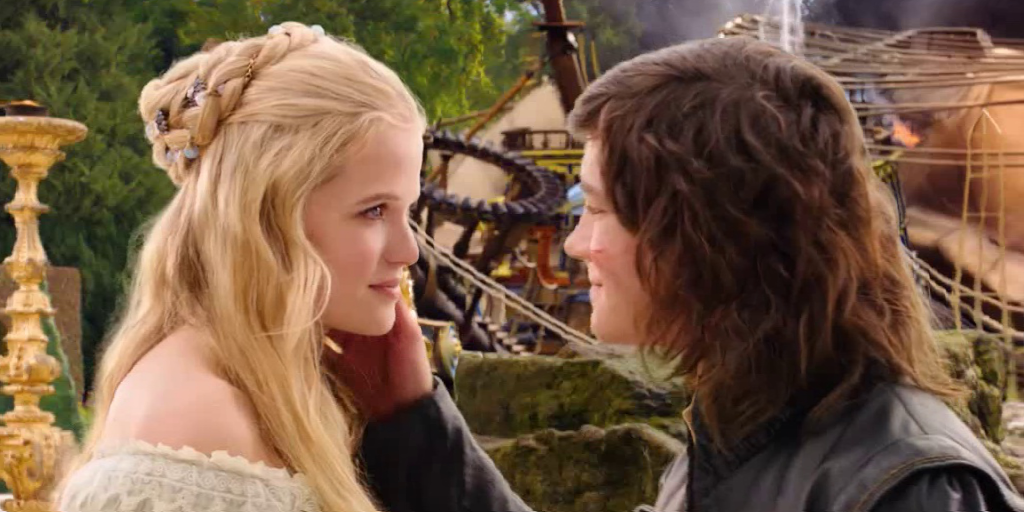}
\includegraphics[width=\linewidth]{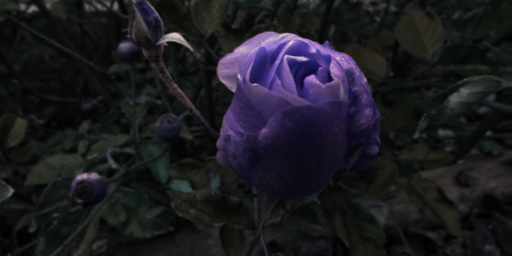}
\endminipage
\hfill
\minipage{0.32\linewidth}
\centering
\begin{subfigure}{0.02\linewidth}
\end{subfigure}
\includegraphics[width=\linewidth]{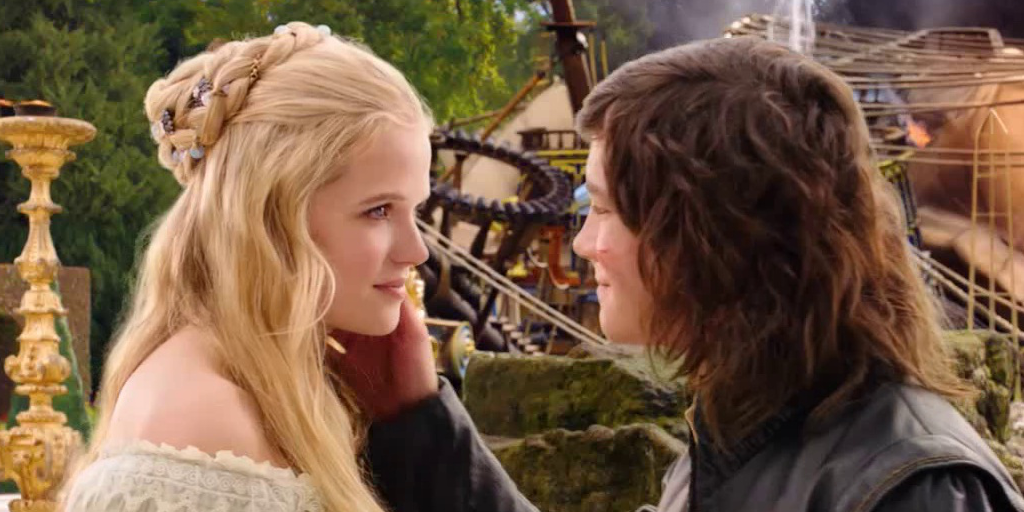}
\includegraphics[width=\linewidth]{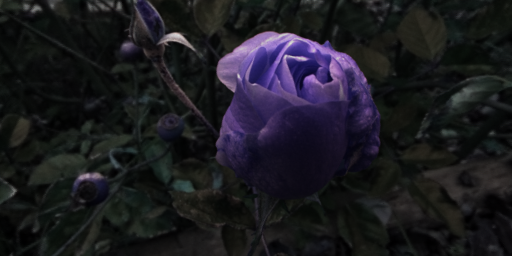}
\endminipage
\hfill
\minipage{0.32\linewidth}
\centering
\begin{subfigure}{0.02\linewidth}
\end{subfigure}
\includegraphics[width=\linewidth]{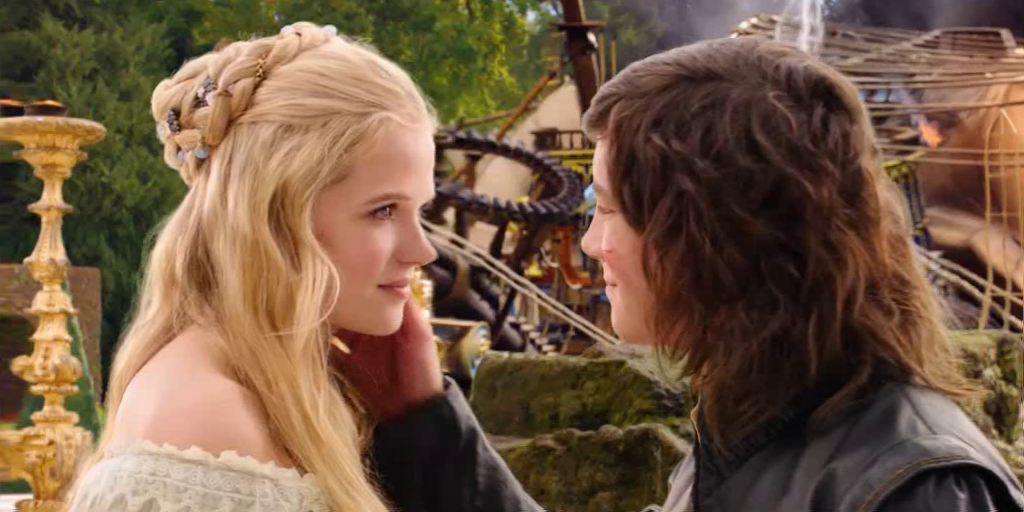}
\includegraphics[width=\linewidth]{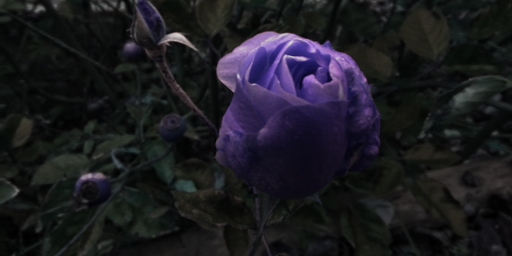}
\endminipage

\caption{Images produced when the network is applied on other datasets (Hollywood \cite{hadfield13} and Flowers \cite{srinivasan17}). From left to right: left image produced, input image, right image produced.}
\label{other_datasets_hollyflowers}
\end{figure*}

\subsection{Limits of the approach}

\begin{figure*}[!htb]
\centering
\minipage{0.19\linewidth}
\centering
\begin{subfigure}{0.02\linewidth}
\caption{}
\end{subfigure}
\includegraphics[width=\linewidth]{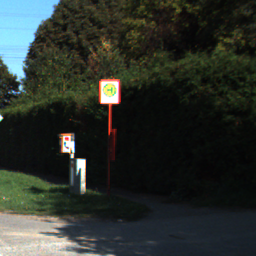}
\includegraphics[width=\linewidth]{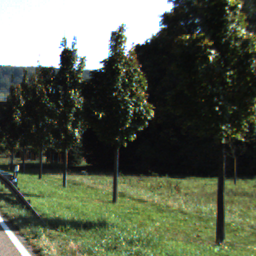}
\endminipage
\hfill
\minipage{0.19\linewidth}
\centering
\begin{subfigure}{0.02\linewidth}
\caption{}
\end{subfigure}
\includegraphics[width=\linewidth]{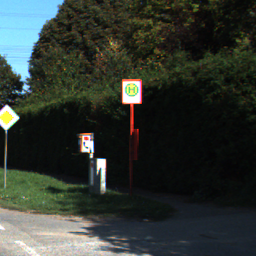}
\includegraphics[width=\linewidth]{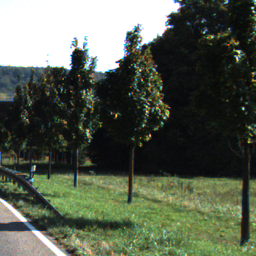}
\endminipage
\hfill
\minipage{0.19\linewidth}
\centering
\begin{subfigure}{0.02\linewidth}
\caption{}
\end{subfigure}
\includegraphics[width=\linewidth]{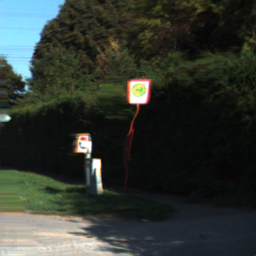}
\includegraphics[width=\linewidth]{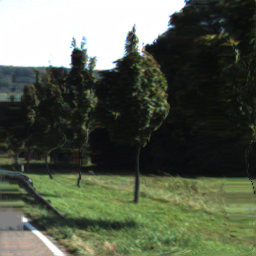}
\endminipage
\hfill
\minipage{0.19\linewidth}
\centering
\begin{subfigure}{0.02\linewidth}
\caption{}
\end{subfigure}
\includegraphics[width=\linewidth]{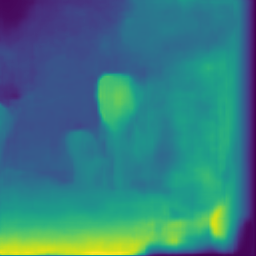}
\includegraphics[width=\linewidth]{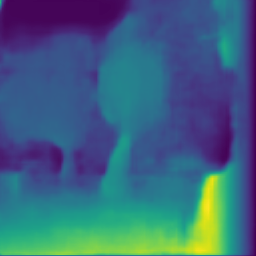}
\endminipage
\hfill
\minipage{0.19\linewidth}
\centering
\begin{subfigure}{0.02\linewidth}
\caption{}
\end{subfigure}
\includegraphics[width=\linewidth]{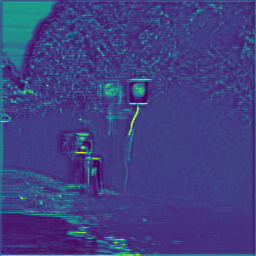}
\includegraphics[width=\linewidth]{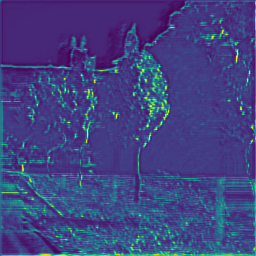}
\endminipage

\caption{Failure cases for our approach. \textbf{a)} Input image. \textbf{b)} Ground truth image. \textbf{c)} Network prediction. \textbf{d)} Estimated disparity map. \textbf{e)} Confidence map.}
\label{failure_cases}
\end{figure*}

If the approach is able to perform convincing view synthesis in most cases, limits have to be highlighted. Since the geometrical refiners are all based on the color gradient, it means that they notably have a hard time segmenting unusual structures for which the color greatly varies (see figure \ref{failure_cases} a)). It is noteworthy, though, that even if the network is not able to process this region correctly, it still classifies it as a low-confidence region.

As a whole, we can note that the method also encounters difficulties when working on thin structures in front of a background with a strong color gradient. In most cases, we also notice that the occluded regions at the border of the image are filled essentially by extending the boundary pixels, and that the algorithm is not capable of coming up with something more sophisticated (see figure \ref{failure_cases} b)). 

\subsection{Conclusion}

We have presented in this paper a supervised CNN-based approach able to perform monocular view synthesis. The MobileNet based-encoder allows us to obtain a good disparity-based prediction with a low number of parameters. This prediction is then refined in regions where artifacts are still present, occluded areas and mispredicted parts using the Refiner. The network is also able to estimate the confidence that it has in its own disparity-based prediction, and is able to identify the structures that it has not predicted correctly. The method outperforms state-of-the-art approaches metric-wise and visually in the domain of monocular view generation on the KITTI dataset. 

\newpage

\bibliographystyle{IEEEtran}
\bibliography{article}

\begin{IEEEbiography}[{\includegraphics[width=1in, height=1.25in,clip,keepaspectratio]{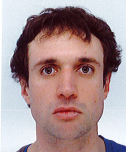}}]{Simon Evain}
Simon Evain is a PhD student at INRIA Rennes. He graduated from École des Ponts ParisTech in 2017, and he holds a Research Master's Degree in computer vision and machine learning from ENS Cachan (MVA). Since 2017, he has been working as a PhD student with University of Rennes 1, under the supervision of Christine Guillemot. His research interests include deep learning, computer vision as well as light field and image processing.
\end{IEEEbiography}

\begin{IEEEbiography}[{\includegraphics[width=1in,height=1.25in,clip,keepaspectratio]{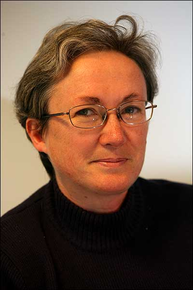}}]{Christine Guillemot}

Christine Guillemot, IEEE fellow, is Director of Research at INRIA. She holds a Ph.D. degree from ENST (Ecole Nationale Superieure des Telecommunications) Paris, and an Habilitation for Research Direction from the University of Rennes. From 1985 to Oct. 1997, she has been with FRANCE TELECOM, where she has been involved in various projects in the area of image and video coding and processing for TV, HDTV and multimedia. From Jan. 1990 to mid 1991, she has worked at Bellcore, NJ, USA, as a visiting scientist. Her research interests are signal and image processing, and computer vision.

She has served as Associate Editor for IEEE Trans. on Image Processing (from 2000 to 2003, and from 2014-2016), for IEEE Trans. on Circuits and Systems for Video Technology (from 2004 to 2006), and for IEEE Trans. on Signal Processing (2007-2009). She has served as senior member of the editorial board of the IEEE journal on selected topics in signal processing (2013-2015) and is currently senior area editor of IEEE Trans. on Image Processing.  \end{IEEEbiography}


\end{document}


\section*{Exact architecture of the network}

\begin{table}[!h]
\centering
\begin{tabular}{|c|c|c|c|}
\hline
\textbf{Number} & \textbf{Type - stride} & \textbf{Number of filters} & \textbf{Kernel size}\\
\hline
 1 & Conv - 2 & 32 & 3*3  \\
 \hline
 2 & Depthwise conv - 1 & & 3*3 \\
 \hline
 3 & Conv - 1 & 64 & 1*1 \\
 \hline
 4 & Depthwise conv - 2 & & 3*3 \\
 \hline
 5 & Conv - 1 & 128 & 1*1 \\
 \hline
 6 & Depthwise conv - 1 & & 3*3 \\
 \hline
 7 & Conv - 1 & 128 & 1*1 \\
 \hline
 8 & Depthwise conv - 2 & & 3*3 \\
 \hline
 9 & Conv - 1 & 256 & 1*1 \\
 \hline
 10 & Depthwise conv - 1 & & 3*3 \\
 \hline
 11 & Conv - 1 & 256 & 1*1 \\
 \hline
 12 & Depthwise conv - 2 & & 3*3 \\
 \hline
 13 & Conv - 1 & 512 & 1*1 \\
 \hline
 14-23 & 5* & & \\
  & Depthwise conv - 1 & & 3*3 \\
  & Conv - 1 & 512 & 1*1 \\
 \hline
 24 & Depthwise conv - 2 & & 3*3 \\
 \hline
 25 & Conv - 1 & 1024 & 1*1 \\
 \hline
 26 & Depthwise conv - 2 & & 3*3 \\
 \hline
 27 & Conv - 1 & 1024 & 1*1 \\
 \hline
\end{tabular}
\caption{Complete architecture of the feature extractor (for both branches). All layers have a relu6 activation function.}
\label{feature_extractor}
\end{table}

\begin{table}[]
\centering
\begin{tabular}{|c|c|c|c|}
\hline
\textbf{Number} &\textbf{Type - stride} & \textbf{Number of filters} & \textbf{Kernel size}\\
\hline
28 & Depthwise conv - 1 & & 3*3 \\
 \hline
29 &  Conv - 1 & 512 & 1*1 \\
\hline
30 & UpSampling - 2 & & \\
\hline
31 & Concatenate (to layer 24) & &\\
 \hline
32 & Depthwise conv - 1 & & 3*3 \\
 \hline
33 &  Conv - 1 & 512 & 1*1 \\
\hline
34 & UpSampling - 2 & & \\
\hline
35 & Concatenate (to layer 12) & &\\
\hline
36 & Depthwise conv - 1 & & 3*3 \\
 \hline
37 &  Conv - 1 & 256 & 1*1 \\
\hline
38 & UpSampling - 2 & & \\
\hline
39 & Concatenate (to layer 8) & &\\
\hline
40 & Depthwise conv - 1 & & 3*3 \\
 \hline
41 &  Conv - 1 & 128 & 1*1 \\
\hline
42 & UpSampling - 2 & & \\
\hline
43 & Concatenate (to layer 4) & &\\
\hline
44 & Depthwise conv - 1 & & 3*3 \\
 \hline
45 &  Conv - 1 & 64 & 1*1 \\
\hline
46 & UpSampling - 2 & & \\
\hline
47 & Conv - 1 & 1 & 2*2 \\
\hline
\end{tabular}
\caption{Complete architecture of the disparity estimator (for one branch). All layers (except for layer 47, with relu) have relu6 as an activation function.}
\label{disparity_estimator}
\end{table}

\begin{table}[]
\centering
\begin{tabular}{|c|c|c|c|}
\hline
\textbf{Number} &\textbf{Type - stride} & \textbf{Number of filters} & \textbf{Kernel size}\\
\hline
48 & Conv - 1 & 64 & 3*3 \\
\hline
49 & Conv - 1 & 64 & 3*3 \\
\hline
50 & Conv - 1 & 64 & 3*3 \\
\hline
51 & Conv - 1 & 64 & 3*3 \\
\hline
52 & Conv - 1 & 64 & 3*3 \\
\hline
53 & Conv - 1 & 64 & 3*3 \\
\hline
54 & Conv - 1 & 64 & 3*3 \\
\hline
55 & Conv - 1 & 3 & 3*3\\
\hline
\end{tabular}
\caption{Complete architecture of the refiner (for one branch). All layers (except for layer 55, with no activation function) have a relu activation function.}
\label{refiner}
\end{table}

\begin{table}
\centering
\begin{tabular}{|c|c|c|c|}
\hline
\textbf{Number} &\textbf{Type - stride} & \textbf{Number of filters} & \textbf{Kernel size}\\
\hline
56 & Conv - 1 & 32 & 3*3 \\
\hline
57 & Conv - 1 & 32 & 3*3 \\
\hline
58 & Conv - 1 & 32 & 3*3 \\
\hline
59 & Conv - 1 & 32 & 3*3 \\
\hline
60 & Conv - 1 & 32 & 3*3 \\
\hline
\end{tabular}
\caption{Complete architecture of the Confidence-Based Merger (for one branch). All layers (except for layer 60, with sigmoid) have a relu activation function.}
\label{confidence}
\end{table}